\documentclass[10pt,twocolumn,letterpaper]{article}

\usepackage{iccv}
\usepackage{times}
\usepackage{epsfig}
\usepackage{graphicx}
\usepackage{amsmath}
\usepackage{amssymb}

\usepackage[table]{xcolor}
\usepackage{arydshln}
\usepackage{multirow}
\usepackage{booktabs}
\usepackage{subfigure}
\usepackage[ruled,linesnumbered]{algorithm2e}
\usepackage{paralist}
\usepackage[accsupp]{axessibility}

\usepackage[pagebackref=true,breaklinks=true,letterpaper=true,colorlinks,bookmarks=false]{hyperref}

\iccvfinalcopy 

\def\iccvPaperID{1626} 

\ificcvfinal\fi

\begin{document}

\title{Contrastive Pseudo Learning for Open-World DeepFake Attribution}

\author{
Zhimin Sun$^{1,2,3}$\footnotemark[1]
\quad Shen Chen$^2$\footnotemark[1]
\quad Taiping Yao$^2$
\quad Bangjie Yin$^2$ \\
\quad Ran Yi$^1$\footnotemark[2]
\quad Shouhong Ding$^2$\footnotemark[2]
\quad Lizhuang Ma$^1$\\
$^1$Shanghai Jiao Tong University
\quad $^2$Tencent YouTu Lab \\
$^3$Shanghai Key Laboratory of Computer Software Testing \& Evaluating
}

\maketitle
\ificcvfinal\thispagestyle{empty}\fi

{
  \renewcommand{\thefootnote}%
    {\fnsymbol{footnote}}
  \footnotetext[1]{Equal contribution. This work was done when Zhimin Sun was a research intern at Tencent YouTu Lab.}
  \footnotetext[2]{Corresponding authors.}
  \footnotetext[3]{Code at: \url{https://github.com/TencentYoutuResearch/OpenWorld-DeepFakeAttribution}}
}


\begin{abstract}
\vspace{-0.25cm}
The challenge in sourcing attribution for forgery faces has gained widespread attention due to the rapid development of generative techniques.
While many recent works have taken essential steps on GAN-generated faces, more threatening attacks related to identity
swapping or expression transferring are still overlooked. And the forgery traces hidden in unknown attacks from the open-world unlabeled faces still remain under-explored.
To push the related frontier research, we introduce a new benchmark called Open-World DeepFake Attribution (OW-DFA), which aims to evaluate attribution performance against various types of fake faces under open-world scenarios.
Meanwhile, we propose a novel framework named Contrastive Pseudo Learning (CPL) for the OW-DFA task through 1) introducing a Global-Local Voting module to guide the feature alignment of forged faces with different manipulated regions, 2) designing a Confidence-based Soft Pseudo-label strategy to mitigate the pseudo-noise caused by similar methods in unlabeled set. In addition, we extend the CPL framework with a multi-stage paradigm that leverages pre-train technique and iterative learning to further enhance traceability performance.
Extensive experiments verify the superiority of our proposed method on the OW-DFA and also demonstrate the interpretability of deepfake attribution task and its impact on improving the security of deepfake detection area.
\end{abstract}

\vspace{-0.8cm}


\section{Introduction}

\begin{figure}[t]
    \centering
    \includegraphics[width=\columnwidth]{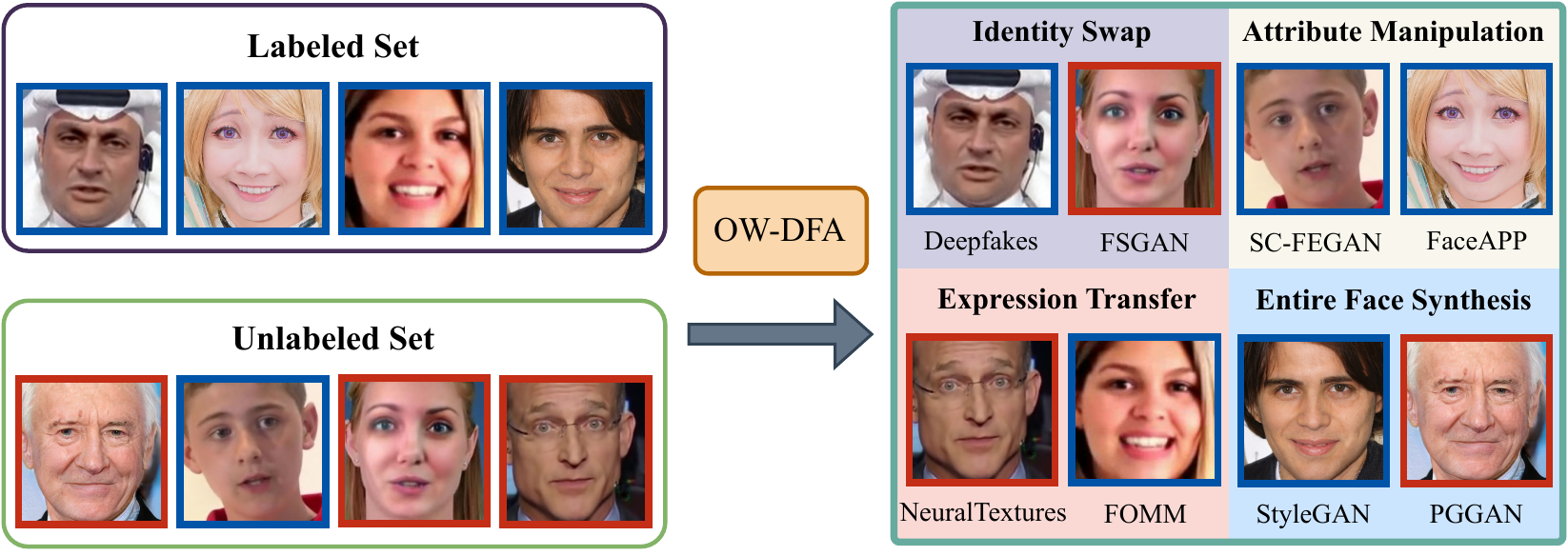}
    \caption{In the OW-DFA setting, the unlabeled dataset may contain attacks that have never been encountered in the labeled set. A feasible model should attribute the known attacks (images with blue border) and assign the unknown attacks (images with red border) to novel classes simultaneously.}
    \label{fig:setting}
    \vspace{-0.5cm}
\end{figure}

With the rapid development of generative technologies such as Deepfakes~\cite{deepfakes}, the malicious usage of fake content on social media has raised public concerns about face security and privacy.
Dedicated research efforts~\cite{haliassos2021lips,shiohara2022detecting,wang2020cnn} have been made in the real/fake detection task in recent years.
Nonetheless, with its distinctive merits, DeepFake Attribution (DFA), known as identifying the source model of fake faces, has also significantly drawn widespread attentions~\cite{yang2022deepfake,yu2021artificial,guarnera2022exploitation}.
On the one hand, DFA can be used for legal proceedings and provide interpretability to human beings, \textit{i.e.}``why the face is fake.''
On the other hand, with the nature of learning enhanced representation for different attacking types, DFA is also effective to boost the deepfake detection performance~\cite{he2021forgerynet,dong2022explaining}.

Early approaches of sourcing attribution~\cite{yang2022deepfake,yu2021artificial,guarnera2022exploitation} mostly focus on the GAN-generated images rather than the more realistic and threatening attacks related to identity swapping or expression transferring.
Meanwhile, most of them assume a closed scenario where the training set and test set share the same category distributions, which is not applicable to open-world scenarios since new types of forgery attacks emerge immensely.
To this end, we introduce a new benchmark, Open-World DeepFake Attribution (OW-DFA), as shown in Figure~\ref{fig:setting}. The OW-DFA benchmark consists of a labeled training dataset and an unlabeled dataset. The labeled dataset contains samples from known classes, while the unlabeled dataset includes samples from both known and unknown classes. More importantly, OW-DFA considers nearly 20 challenging and realistic forgery methods, including 4 widely-used forgery types, namely \textit{identity swap}~\cite{DeepFaceLab, deepfakes}, \textit{expression transfer}~\cite{NeuralTextures,fomm}, \textit{attribute manipulation}~\cite{choi2018stargan,choi2020starganv2} and \textit{entire face synthesis}~\cite{karras2018progressive,Karras2019stylegan2}. The main challenge of OW-DFA is how to utilize unlabeled data in open-world scenes to improve the attribution performance for both known and unknown forged faces.

The OW-DFA is fundamentally different but closely related to Open-World Semi-Supervised Learning (OW-SSL), where some OW-SSL methods~\cite{han2020automatically,cao2022openworld,guo2022robust} demonstrate effectiveness in learning unknown categories through contrastive learning or pseudo-labeling strategies.
However, since all classes of OW-DFA are face data whose disparity information relies on fine-grained forgery traces~\cite{chen2021local,zhao2020learning}, these OW-SSL methods that only focus on global information will be limited in attributing unknown forged faces. 
Moreover, Open-world GAN~\cite{girish2021towards} discovers and refines unseen GANs with iterative algorithms.
However, the fingerprint assumption relied on may not hold in the fake faces generated by non-GAN methods. Without a semi-supervised learning strategy, the features extracted by this model for different unknown attacks lack distinguishability.

In this paper, we propose a novel framework named Contrastive Pseudo Learning (CPL), which addresses the above issues from two perspectives:
1) We introduce a Global-Local Voting (GLV) module that guides inter-sample feature alignment by extracting both global and local information and adaptively highlights different manipulated regions through a spatially enhancing mechanism. By combining global and local similarity, we can filter and group together samples of the same attack type.
2) Besides the inter-sample relation, we also leverage the intra-sample information to enhance the class compactness using the pseudo-labeling technique. A Confidence-based Soft Pseudo-labeling (CSP) mechanism is proposed to mitigate the pseudo-noise induced by similar novel attack methods.
Moreover, previous research~\cite{han2020automatically,vaze2022generalized} has demonstrated the efficacy of pre-training techniques and iterative learning, so we extend the CPL framework with a multi-stage paradigm to further improve the attribution performance.
Finally, extensive experimental results verify the superiority of our method on the OW-DFA benchmark. We also demonstrate the interpretability of the deepfake attribution task and its impact on improving the security of the deepfake detection area.

We summarize our contributions as follows:

(1) We present a new benchmark called Open-World DeepFake Attribution (OW-DFA), which aims to evaluate attribution performance against various types of fake faces under open-world scenarios.

(2) We propose a novel Contrastive Pseudo Learning (CPL) framework for OW-DFA task through 1) a Global-Local Voting module to guide the feature alignment of forged faces with different manipulated regions, 2) a Confidence-based Soft Pseudo-labeling strategy to mitigate pseudo-noise caused by similar methods in unlabeled set.

(3) Comprehensive experiments and visualization results demonstrate that our method achieves SOTA performance on OW-DFA. We also show that combining the deepfake attribution task with the deepfake detection task leads to better interpretability and face security.

\begin{table*}[th]
\begin{center}
\caption{List of methods and corresponding datasets utilized in OW-DFA. Protocol 1 encompasses 20 challenging forgery techniques, with forgery types ranging from identity swap, expression transfer, attribute manipulation and entire face synthesis. The primary objective of Protocol 1 is to enhance the attribution of forgery attacks. Protocol 2 combines the forgery techniques from Protocol 1 with real faces to create a realistic open-set mixed attribution scenario that mimics real-life situations.}
\label{tab:dataset_real}
\resizebox{\textwidth}{!}{%
\begin{tabular}{@{}cllccccc@{}}
\toprule
\textbf{Face Type} &
  \textbf{Labeled Sets} &
  \textbf{Unlabeled Sets} &
  \textbf{Source Dataset} &
  \textbf{Method} &
  \textbf{Tag} &
  \textbf{Labeled \#} &
  \textbf{Unlabeled \#} \\ \midrule
\multirow{5}{*}[-2pt]{Identity Swap} &
  \multirow{5}{*}[-2pt]{\begin{tabular}[c]{@{}l@{}}Deepfakes~\cite{deepfakes}\\ DeepFaceLab~\cite{DeepFaceLab}\end{tabular}} &
  \multirow{5}{*}[-2pt]{\begin{tabular}[c]{@{}l@{}}Deepfakes\\ DeepFaceLab\\ FaceSwap~\cite{faceswap}\\ FaceShifter~\cite{li2019faceshifter}\\ FSGAN~\cite{nirkin2019fsgan}\end{tabular}} &
  \multirow{2}{*}{FaceForensics++~\cite{rossler2019ff++}} &
  Deepfakes & Known & 1500 & 500 \\
 &  &  & & FaceSwap & Novel & - & 1500 \\ \cmidrule(lr){4-8}
 &  &  & \multirow{3}{*}{ForgeryNet~\cite{he2021forgerynet}} & DeepFaceLab & Known & 1500 & 500 \\
 &  &  & & FaceShifter & Novel & - & 1500 \\
 &  &  & & FSGAN & Novel & - & 1500\\ \midrule
\multirow{5}{*}[-2pt]{Expression Transfer} &
  \multirow{5}{*}[-2pt]{\begin{tabular}[c]{@{}l@{}}Face2Face~\cite{face2face}\\ FOMM~\cite{fomm}\end{tabular}} &
  \multirow{5}{*}[-2pt]{\begin{tabular}[c]{@{}l@{}}Face2Face\\ FOMM \\ NeuralTextures~\cite{NeuralTextures}\\ Talking-Head-Video~\cite{zhang2021text2video}\\ ATVG-Net~\cite{chen2019hierarchical}\end{tabular}} &
  \multirow{2}{*}{FaceForensics++} & Face2Face & Known & 1500 & 500  \\
 &  &  & & NeuralTextures     & Novel & -    & 1500 \\ \cmidrule(lr){4-8}
 &  &  & \multirow{3}{*}{ForgeryNet} & FOMM               & Known & 1500& 500   \\
 &  &  & & ATVG-Net           & Novel & -    & 1500 \\
 &  &  & & Talking-Head-Video & Novel & -    & 1500 \\ \midrule
\multirow{5}{*}[-2pt]{Attribute Manipulation} &
  \multirow{5}{*}[-2pt]{\begin{tabular}[c]{@{}l@{}}MaskGAN~\cite{CelebAMaskGAN-HQ}\\ FaceAPP~\cite{faceapp}\end{tabular}} &
  \multirow{5}{*}[-2pt]{\begin{tabular}[c]{@{}l@{}}MaskGAN\\ FaceAPP\\ StarGAN2~\cite{choi2020starganv2}\\ SC-FEGAN~\cite{Jo_2019_ICCV}\\ StarGAN~\cite{choi2018stargan}\end{tabular}} &
  \multirow{3}{*}{ForgeryNet} & MaskGAN & Known & 1500 & 500 \\
 &  &  & & StarGAN2 & Novel & - & 1500 \\
 &  &  & & SC-FEGAN & Novel & - & 1500 \\ \cmidrule(lr){4-8}
 &  &  & \multirow{2}{*}{DFFD~\cite{dang2020detection}} & FaceAPP & Known & 1500 & 500 \\
 &  &  & & StarGAN & Novel & - & 1500 \\ \midrule
\multirow{5}{*}[-4pt]{Entire Face Synthesis} &
  \multirow{5}{*}[-4pt]{\begin{tabular}[c]{@{}l@{}}StyleGAN~\cite{karras2019style}\\ CycleGAN~\cite{CycleGAN2017}\end{tabular}} &
  \multirow{5}{*}[-4pt]{\begin{tabular}[c]{@{}l@{}}StyleGAN\\ CycleGAN\\ PGGAN~\cite{karras2018progressive}\\ StyleGAN2~\cite{Karras2019stylegan2}\end{tabular}} &
  ForgeryNet & StyleGAN2 & Novel & - & 1500\\ \cmidrule(lr){4-8}
 &  &  & \multirow{2}{*}{DFFD} & StyleGAN & Known & 1500 & 500  \\
 &  &  & & PGGAN & Novel &  -   & 1500 \\ \cmidrule(lr){4-8}
 &  &  & \multirow{2}{*}{ForgeryNIR~\cite{wang2022forgerynir}} & CycleGAN           & Known & 1500& 500    \\ 
 &  &  & & StyleGAN2 & Novel &  -   & 1500 \\ \midrule 
\multirow{2}{*}[-2pt]{Real Face} & \multirow{2}{*}[-2pt]{\begin{tabular}[c]{@{}l@{}} Youtube-Real~\cite{rossler2019ff++}\end{tabular}} & \multirow{2}{*}[-2pt]{\begin{tabular}[c]{@{}l@{}} Celeb-Real~\cite{li2020celeb}\end{tabular}} & FaceForensics++ & Youtube-Real & Known & 15000 & 5000 \\ \cmidrule(lr){4-8}
 & & & CelebDFv2~\cite{li2020celeb} & Celeb-Real & Novel & - & 5000 \\ \bottomrule
\end{tabular}%
}
\end{center}
\end{table*}

\section{Related Works}

\subsection{DeepFake Attribution}
A plethora of works~\cite{chollet2017xception,tan2019efficientnet,qian2020thinking,zhao2021multi,chen2022shape,cao2022end,sun2022dual,sun2022information,gu2022exploiting,gu2021spatiotemporal,gu2022region,gu2022hierarchical,gu2022delving} for the real/fake detection task have been proposed in recent years. However, the generalization performance on novel attacks is still limited.
As fake faces become visually realistic and need to be interpreted in legal proceedings, attribution of the source model of fake faces has gained widespread attention.
Most existing works~\cite{yang2022deepfake,yu2019attributing,yu2021artificial,guarnera2022exploitation} focus only on the problem of attributing GAN models, and a common strategy is to use the fingerprints of different GAN models to attribute those generated images. However, they only consider the close-world scenario where the training and test sets have the same category distribution. Such an assumption is not applicable to open-world scenarios since novel forgeries emerge greatly.
The most relevant method, Open-world GAN~\cite{girish2021towards}, proposes an iterative algorithm to discover and refine unseen GANs in an open-world scenario. Although it has made some progress in open-world scenarios, the features extracted by this model for unknown attacks lack discriminability without proper use of unlabeled data.
To this end, we propose a new benchmark OW-DFA, which contains samples from both known and unknown classes and then utilizes more challenging and realistic forgery methods. Furthermore, with the proposed CPL framework, we significantly boost the performance of deepfake attribution under the OW-DFA setup.

\subsection{Open-World Semi-Supervised Learning}
Open-World Semi-Supervised Learning (OW-SSL)~\cite{cao2022openworld} aims to leverage both labeled and unlabeled data in an open-world scenario, where novel classes may exist in the unlabeled datasets. Existing OW-SSL methods~\cite{cao2022openworld,rizve2022openldn,guo2022robust} bring instances of the same class in unlabeled datasets closer together based on inter-sample similarity.
And assigning pseudo-labels to the high-confidence samples is another common technique~\cite{sohn2020fixmatch,zhang2021flexmatch,wang2022freematch,rizve2022openldn,rizve2022towards}.
Despite the promising performance of these methods on OW-DFA tasks, they still face some challenges. First, existing methods~\cite{cao2022openworld,rizve2022openldn} mainly focus on the global similarity of samples, neglecting the local consistency of forged face images that may indicate tampering.
To address this, we propose a Global-Local Voting module that matches samples more accurately by considering both global and local features of facial attacks.
Second, the threshold-based pseudo-labeling strategy~\cite{zhang2021flexmatch,wang2022freematch} can only handle samples with deterministic labels.
To address this, we propose a probability-based pseudo-labeling strategy that imposes additional constraints on samples with low confidence.

\section{Open-World DeepFake Attribution}
In this section, we first present the definition of the Open-World DeepFake Attribution (OW-DFA) task with labeled and unlabeled sets, known and novel categories, and the corresponding notation.
We then list the dataset composition of OW-DFA and propose two challenging protocols.
More preprocessing details for each dataset are provided in
Appendix Sec.~\ref{sec:prepro}.

\subsection{Definition}
The Open-World DeepFake Attribution (OW-DFA) task consists of a labeled set $\mathcal{D}_l=\left\{\left(x_i, y_i\right)\right\}_{i=1}^n$ and an unlabeled set $\mathcal{D}_u=\left\{\left(x_i\right)\right\}_{i=1}^m$.
We denote the classes in the labeled set as $\mathcal{C}_L$, and those in the unlabeled set as $\mathcal{C}_U$, where $\mathcal{C}_L$ contains only known categories, while $\mathcal{C}_U$ covers both known and novel categories, \textit{i.e.}, $\mathcal{C}_L \cap \mathcal{C}_U \neq \emptyset$ and $\mathcal{C}_L \ne \mathcal{C}_U$.
We denote the known class as $\mathcal{C}_K = \mathcal{C}_L \cap \mathcal{C}_U$ and the novel class as $\mathcal{C}_N = \mathcal{C}_U \backslash \mathcal{C}_L$.
The goal of OW-DFA task is utilizing both labeled sets $\mathcal{D}_l$ and unlabeled sets $\mathcal{D}_u$ to learn a feature extractor $\phi(\cdot)$ and a classifier $\sigma(\cdot)$, which can recognize source models for various faces types.
Unlike previous work~\cite{girish2021towards,guarnera2022exploitation,yang2022deepfake,yu2019attributing,yu2021artificial} that only considered GAN-generated images, OW-DFA also includes more threatening attacks related to identity swap or expression transfer.

\subsection{Protocols}\label{sec:exp}
We create the OW-DFA benchmark based on several deepfake datasets, including FF++~\cite{rossler2019ff++}, CelebDF~\cite{li2020celeb}, ForgeryNet~\cite{he2021forgerynet}, DFFD~\cite{dang2020detection} and ForgeryNIR~\cite{wang2022forgerynir}.
These datasets are widely used in the deepfake detection task with large-scale data and various types of forged faces, which can be roughly divided into 5 face types: \textit{identity swap}, \textit{expression transfer}, \textit{attribute manipulation}, \textit{entire face synthesis} and \textit{real face}.

As can be seen in Table~\ref{tab:dataset_real}, we define two protocols for OW-DFA to evaluate the performance in real-world scenarios:
1) \textbf{Protocol-1} aims to evaluate the attribution performance of the forgery method, which includes 20 manipulation methods across 4 mainstream forgery face types: \textit{identity swap}, \textit{expression transfer}, \textit{attribute manipulation} and \textit{entire face synthesis}. Under this setting, all labeled and unlabeled data are fake faces. 
2) \textbf{Protocol-2} includes additional real faces from different domains on top of Protocol-1, taking into account the fact that real faces may appear on social platforms. Specifically, we introduce real faces from the FaceForensics++ and Celeb-DF datasets in labeled sets and unlabeled sets respectively. Compared to each forgery type, the amount of real data is larger to simulate the distribution of faces in real scenes.

\begin{figure*}[t]
\begin{center}
\includegraphics[width=\linewidth]{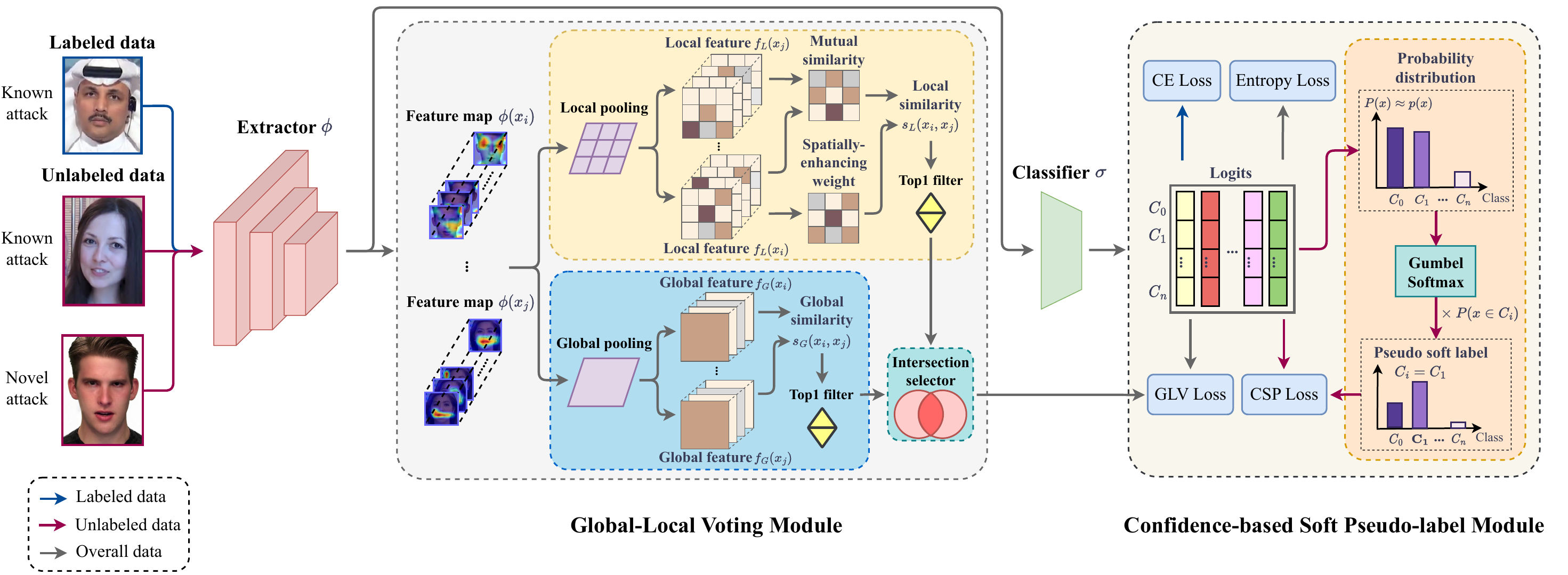}
\end{center}
\caption{Contrastive Pseudo Learning (CPL) framework for Open-World DeepFake Attribution task.}
\label{fig:framework}
\end{figure*}

\section{Contrastive Pseudo Learning Framework}
The key challenge of OW-DFA is to use labeled and unlabelled sets to jointly learn discriminative representations of known and novel attacks.
To this end, we proposed a novel Contrastive Pseudo Learning (CPL) framework, as shown in Figure~\ref{fig:framework}. The CPL framework includes two key components: 1) a Global-Local Voting (GLV) module to guide the feature alignment of different forgery types. 2) a Confidence-based Soft Pseudo-labeling (CSP) module to mitigate the pseudo-noise caused by similar forgery methods in unlabeled sets.
Then we summarise all relevant loss functions.
Finally, we combine the proposed CPL framework with a pretraining technique and iterative learning to further improve the performance under the OW-DFA setup.

\subsection{Global-Local Voting Module}

To facilitate the representation compactness of novel attacks in unlabeled sets, one feasible strategy is the contrast learning~\cite{cao2022openworld,han2020automatically}, which aims to transform the unsupervised clustering problem into a similarity measurement problem.
In particular, given an input face image $x_i$ and label $y^l_i$ for labeled sample, we use $\phi(x_i)$ to extract the corresponding feature map, and a Pooling layer is applied to obtain the global representation, which is formulated as:
\begin{equation}
    \mathit{f}_G(x_i) = \mathit{Pooling}(\phi(x_i); 1 \times 1),
\end{equation}
where $\mathit{f}_G(x_i) \in \mathbb{R}^{d}$, and $d$ denotes the feature dimensions.
For each pair $\left\{ (x_i, x_j): i, j \in (0, \cdots, n + m) \right\}$, the inter-sample relation is measured by the cosine similarity of their global features:
\begin{equation}
    \mathit{s}_G(x_i, x_j) = \frac{\mathit{f}_G(x_i) \cdot \mathit{f}_G(x_j)}{\|\mathit{f}_G(x_i)\|\|\mathit{f}_G(x_j)\|}.
\end{equation}
Given a mini-batch containing both $n$ labeled and $m$ unlabeled samples, we use the above strategy to compute the similarity between each sample $x_i$ and all other samples.
Then we bring $x_i$ closer to its most similar sample $\widetilde{x}_{i}$ by a variant of BCE loss, \ie global relation constraints:
\begin{equation}
\mathcal{L_\mathrm{GR}} = - \frac{1}{n + m} \sum_{x_i \in \mathcal{D}_l \cup \mathcal{D}_u} \log \langle\sigma(\mathit{f}_G(x_i)), \sigma(\mathit{f}_G(\widetilde{x}_{i}))\rangle,
\end{equation} 
where $\sigma$ outputs the probability for each sample.

However, the tempered region varies for different forged types, \textit{e.g.}, the GAN-generated images are forged at every pixel, whereas expression transfer tends to manipulate in the mouth region.
When comparing the similarity of face samples, not considering local fine-grained traces may lead to incorrect contrastive constraints.
Previous works~\cite{chen2021local,zhao2021learning} have shown that integrating global and local information can enhance feature learning. Building upon these findings, we further incorporate local information as well as a voting mechanism to select high-quality pairs. Specifically, we slice the feature map $\phi(x_i)$ for each sample $x_i$ into $q \times q$ regions and the corresponding local representation is obtained as follows:
\begin{equation}
    \mathit{f}_L(x_i) = \mathit{Pooling}(\phi(x_i); q \times q),
\end{equation}
where $\mathit{f}_L(x_i) \in \mathbb{R}^{d \times \mathit{q} \times \mathit{q}}$.
Then we calculate the patch-wise similarity of each sample pair at the same location by cosine similarity:
\begin{equation}
    \mathit{s}^k_{L}(x_i, x_j) = \frac{\mathit{f}^k_L(x_i) \cdot \mathit{f}^k_L(x_j)}{\|\mathit{f}^k_L(x_i)\|\|\mathit{f}^k_L(x_j)\|},
\end{equation}
where $k$ represents the $k$-th patch in $\mathit{f}_L(x_i)$.

Given that manipulated areas may vary across different forged faces, we further introduce a spatially enhancing mechanism to adjust the priority of patch-wise similarities.
MAT~\cite{zhao2021multi} has shown that manipulated areas of forged faces tend to have a higher response, while norm-based analysis~\cite{kobayashi2020attention} demonstrates the effectiveness of $L_2$-norm based attention modules. Inspired by these findings, we use $L_2$-norm to reflect the response of local blocks $f^k_L(x_i)$.
We first calculate the priority weight of \textit{k}-th patch for each sample ${x_i}$ as follows:
\begin{equation}
    \begin{aligned}
    w_i^k = \frac{\|f^k_L(x_i)\|_2}{\sum\limits_{k=1}^{q^2} \|f^k_L(x_i)\|_2}.
    \end{aligned}
\end{equation}
Combining with spatially enhancing weights, the local similarity $s_L(x_i, x_j)$ is obtained:
\begin{equation}
    \begin{aligned}
        s_L(x_i, x_j) = \sum_{k=1}^{q^2} w_i^k \cdot \mathit{s}^k_{L}(x_i, x_j).
    \end{aligned}
\end{equation} 

Next, we propose a voting strategy to take the global and local similarities into consideration.
Given an unlabeled sample $x_i^u$, we can find the two most similar samples $\widetilde{x}_i^u$ and $ \widehat{x}_i^u$ based on Top-1 global similarity $s_G$ and local similarity $s_L$, respectively.
If the results of two Top-1 sample are consistent, \textit{i.e.}, $\widetilde{x}_i^u = \widehat{x}_i^u$, then the pair $(x_i^u, \widetilde{x}_i^u)$ is constrained to be close.
For labeled sample $x_i^l$, we randomly select another sample $\widetilde{x}_i^l$ that belongs to the same class $y_i^l$ in the same batch.
The ultimate loss function for Global-Local Voting module $\mathcal{L_\mathrm{GLV}}$ is formulated as below:
\begin{equation}
    \begin{aligned}
    \mathcal{L_\mathrm{GLV}} =  -\frac{1}{n} & \sum_{x_i \in \mathcal{D}_l} \log \langle\sigma(\mathit{f}_G(x_i^l)), \sigma(\mathit{f}_G(\widetilde{x}_{i}^{l}))\rangle \\
    -\frac{1}{m} & \sum_{x_i \in \mathcal{D}_u} \mathbb{I}\left( \widetilde{x}_i^u = \widehat{x}_i^u \right) \log \langle\sigma(\mathit{f}_G(x_i^u)), \sigma(\mathit{f}_G(\widetilde{x}_{i}^{u}))\rangle.
    \end{aligned}
\end{equation}

\subsection{Confidence-based Soft Pseudo-labeling Module}

\begin{figure}[t]
    \centering
    \subfigure{
    \includegraphics[width=0.305\columnwidth]{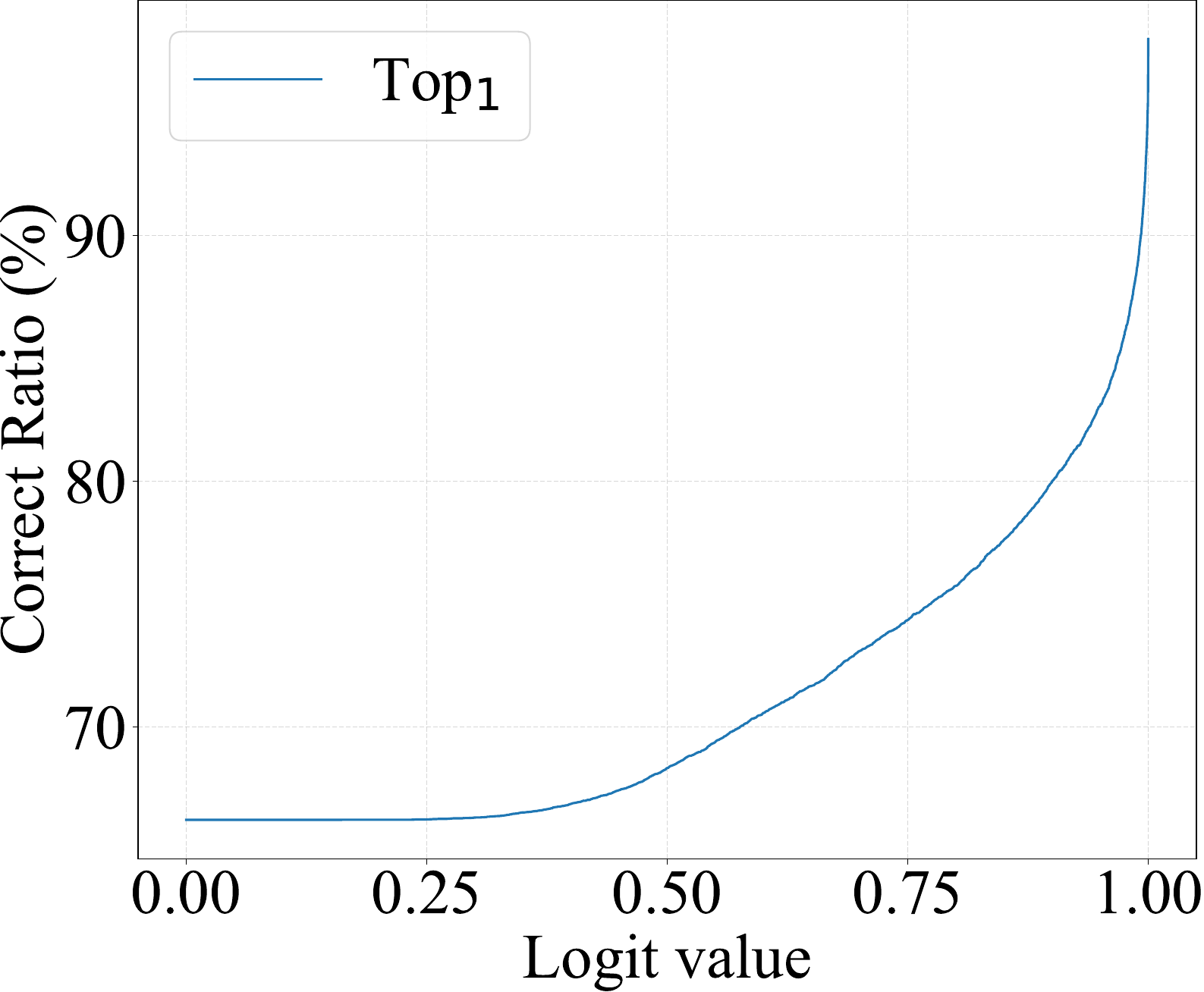}
    }
    \subfigure{
    \includegraphics[width=0.305\columnwidth]{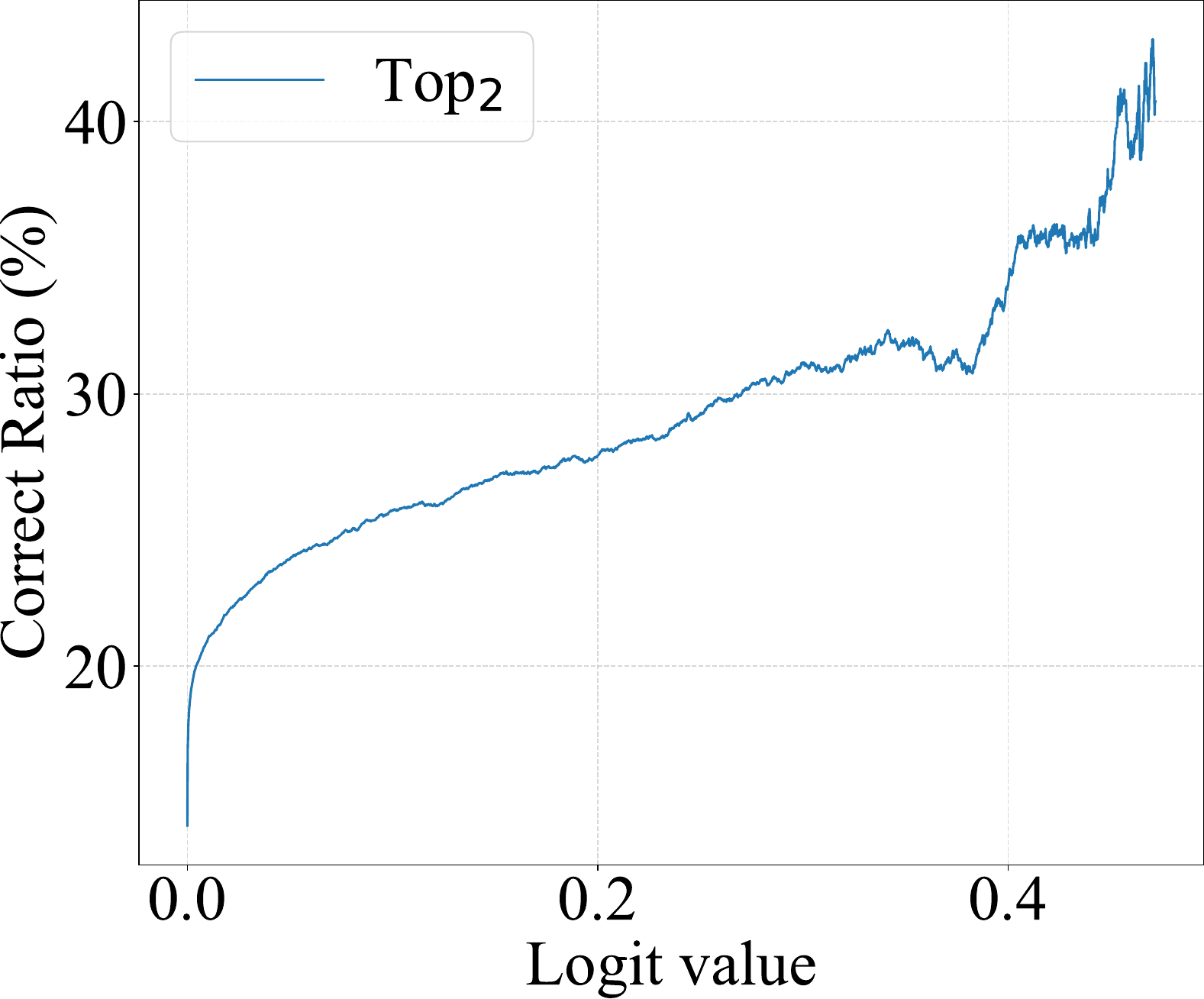}
    }
    \subfigure{
    \includegraphics[width=0.305\columnwidth]{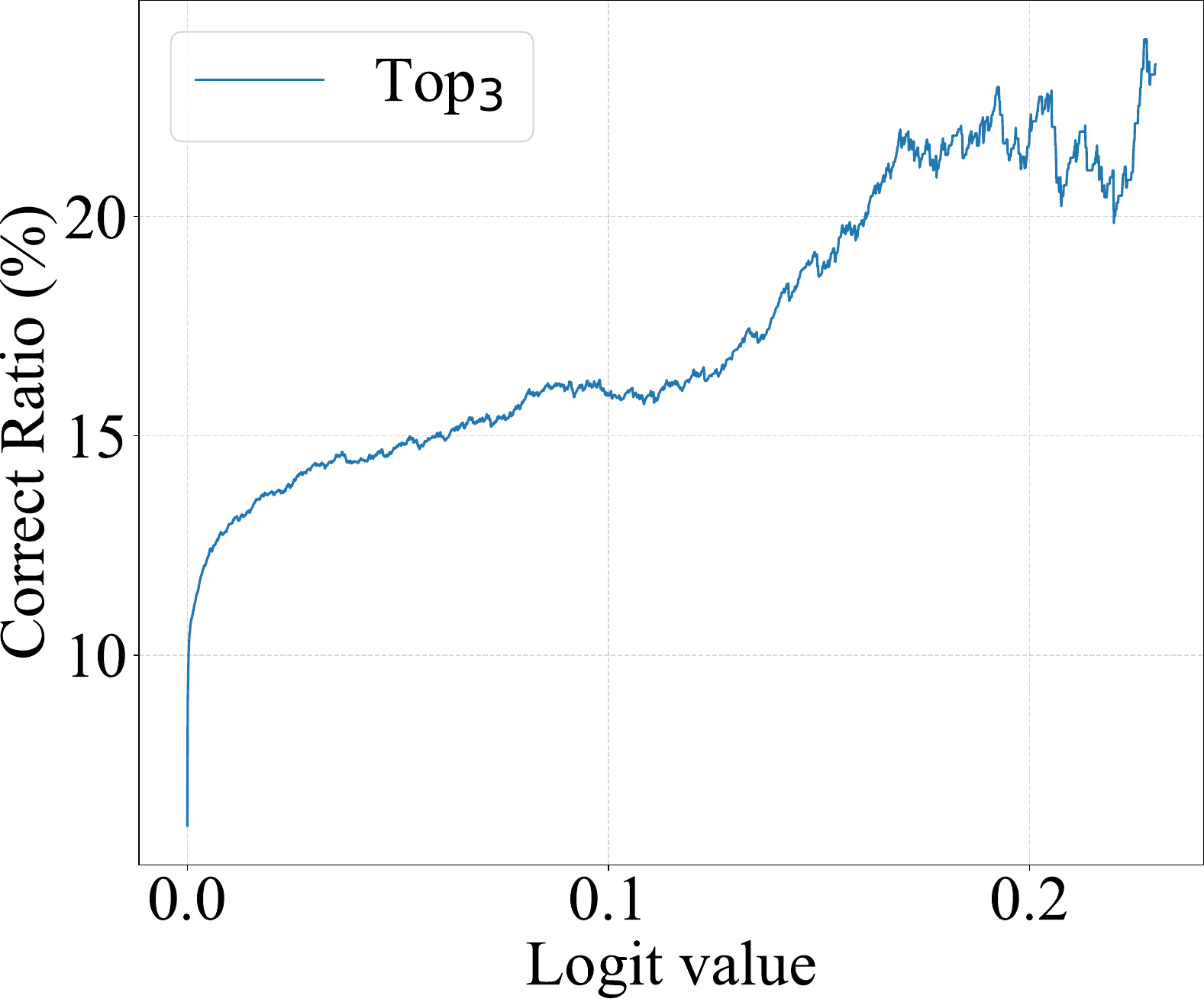}
    }
    \caption{Study of the correlation between the Top-3 value of logits and correct ratio. The results indicate that the second and third predictions still have a high probability of being the correct class.}
    \label{fig:topk}
\end{figure}

With the contrastive learning described above, faces of the same forgery type can be grouped, but some samples with similar manipulated regions may be mixed with other classes without proper supervision.
Pseudo-labeling is a feasible solution that uses the predicted category with the highest probability as classification supervision. However, from the study in Figure~\ref{fig:topk}, we found that the second and the third predictions still have a high probability of being the correct class. 
Therefore, only considering the Top-1 prediction would introduce noisy samples.

Inspired by the study, we propose a Confidence-based Soft Pseudo-labeling module that assigns a pseudo-label for each unlabeled sample based on the output probability of all classes.
For each unlabeled sample $x_i^u$, we first obtain the class probability through $p_i^u = \sigma(\mathit{f}_G(x_i^u))$, where $p_i^u \in \mathbb{R}^{|\mathcal{C}_K \cup \mathcal{C}_N|}$.
Then we introduce the Gumbel Softmax~\cite{jangcategorical} to generate pseudo-label $\widetilde{y}_i^u$ based on the probability $p_i^u$ as follows:
\begin{equation}
    \widetilde{y}_i^u = \mathit{GumbelSoftmax}\left(p_i^u\right).
\end{equation}

We further use the probability of the pseudo-label as a weight to reduce the impact of pseudo-noise when the probability of the assigned pseudo-label is low, and vice versa. The dynamic weight can be calculated through $\lambda_i^u = p_{ic}^{u}$, 
where $c\!=\!\arg \max \widetilde{y}_i^u$.
Finally, we apply soft pseudo-labels of unlabeled data by cross-entropy loss as follows:
\vspace{-0.1cm}
\begin{equation}
    \mathcal{L_\mathrm{CSP}} = -\frac{1}{m} \sum_{x_i \in \mathcal{D}_u} \sum_{c \in \mathcal{C}_K \cup \mathcal{C}_N} \lambda_i^u \cdot \widetilde{y}_{ic}^u \log p_{ic}^u.
\end{equation}

\vspace{-0.2cm}

\subsection{Loss Functions and Multi-stage Paradigm}
Besides the above constraints, we include two loss functions widely used in semi-supervised learning: a cross-entropy loss for labeled data $\mathcal{L_\mathrm{CE}}$, and a regularization term $\mathcal{R}$ to avoid a trivial solution of assigning all instances to the same class, which formulated as follows:
\begin{equation}\label{eq:ce}
\mathcal{L_\mathrm{CE}} = - \frac{1}{n} \sum_{x_i \in \mathcal{D}_l} \sum_{c \in \mathcal{C}_K} y_{ic}^l \log p_{ic}^l,
\end{equation}
\vspace{-0.4cm}
\begin{equation}\label{eq:r}
\mathcal{R} = K L \left( \frac{1}{n+m} \sum_{x_i \in \mathcal{D}_l \cup \mathcal{D}_u} \sigma(\mathit{f}_G(x_i)) \| \mathcal{P}(y)\right) ,
\end{equation} where $p_i^l = \sigma(\mathit{f}_G(x^l_i))$ is class probability and $\mathcal{P}$ denotes a prior probability distribution of labels $y$.
The final loss function is given by: \begin{equation}\label{eq:total_loss}
\mathcal{L} = \mathcal{L_\mathrm{CE}} + \eta_1 \mathcal{L_\mathrm{GLV}} + \eta_2 \mathcal{L_\mathrm{CSP}} + \eta_3 \mathcal{R},
\end{equation}
with hyper-parameters $\eta_1$, $\eta_2$ and $\eta_3$.

Moreover, previous studies~\cite{han2020automatically,vaze2022generalized} have established the effectiveness of pre-training techniques and iterative learning, hence we extend the CPL framework with a multi-stage paradigm, as outlined in Algorithm~\ref{alg:framework}.
In Stage 1, we first pre-train on labeled data to achieve robust performance for known classes.
In Stage 2, we apply the CPL approach on labeled and unlabeled data to discover and enhance the representation of novel attacks.
In Stage 3, we leverage the Semi-Supervised \textit{k}-means algorithm~\cite{vaze2022generalized} to cluster unlabeled samples and assign pseudo-labels based on cluster assignments. We then fine-tune the model with labels in both labeled set $\mathcal{D}_L$ and generated pseudo-labels set $\widetilde{\mathcal{D}}_U$.
With the multi-stage paradigm, we can further improve the attribution performance on the OW-DFA task.
More details are provided in
Appendix Sec.~\ref{sec:msp}.

\newlength{\textfloatsepsave}
\setlength{\textfloatsepsave}{\textfloatsep}
\setlength{\textfloatsep}{0pt}

\begin{algorithm}[t]
	\caption{Multi-stage Paradigm for OW-DFA.}
	\label{alg:framework}
	\KwData{Labeled set $\mathcal{D}_L = {(x_i^l, y_i^l)}_{i=1}^{n}$, Unlabeled set $\mathcal{D}_U = {(x_i^u)}_{i=1}^{m}$.}
	\KwIn{Feature extractor $\phi(\cdot)$, Classifier $\sigma(\cdot)$, Iteration times $T_1, T_2, T_3$.}

	\BlankLine
    Initialize $\phi(\cdot)$ with ImageNet pre-trained weights;

	Initialize $\sigma(\cdot)$ randomly;
    
	\BlankLine
    \leftline{$\triangleright$ Stage 1: Pre-training on labeled-set}
    
    \For{$t$ in \tt{range($T_1$)}}{
    \For{$ (x_i^l, y_i^l) \in \mathcal{D}_L$}{Update $\phi(\cdot)$ and $\sigma(\cdot)$ with  Eq.~\ref{eq:ce};}
    }
	\BlankLine
    \leftline{$\triangleright$ Stage 2: Contrastive Pseudo Learning}
    
    \For{$t$ in \tt{range($T_2$)}}{
    \For{$ (x_i^l, y_i^l) \in \mathcal{D}_L$, $ x_i^u \in \mathcal{D}_U$}{Update $\phi(\cdot)$ and $\sigma(\cdot)$ with Eq.~\ref{eq:total_loss};}
    }
	\BlankLine
    \leftline{$\triangleright$ Stage 3: Iterative Learning}

    $S_L$ = ${(\phi(x_i^l), y_i^l)}_{i=1}^{n}$;
    $S_U$ = ${(\phi(x_i^u))}_{i=1}^{m}$;
    
    $\widetilde{\mathcal{D}}_U$ = Semi-Sup \textit{k}-means($S_L, S_U$);
    
    \For{$t$ in \tt{range($T_3$)}}{
    \For{$ (x_i^l, y_i^l) \in \mathcal{D}_L$ , $(x_i^u, \widetilde{y}_i^u) \in \widetilde{\mathcal{D}}_U$}{Update $\phi(\cdot)$ and $\sigma(\cdot)$ with  Eq.~\ref{eq:ce};}
    }
    \Return $\phi$, $\sigma$
\end{algorithm}

\setlength{\textfloatsep}{\textfloatsepsave}

\section{Experiments}
\textbf{Implementation Details.}
We implement the proposed approach via PyTorch. All the models are trained on $1$ NVIDIA 3090Ti GPU. We use ResNet-50~\cite{he2016deep} pre-trained on ImageNet~\cite{deng2009imagenet} as our feature extractor, and a fully-connected layer as the classifier. We resize the input image to $256 \times 256$, and train the network with Adam~\cite{kingma2014adam} optimizer, a learning rate of $2e^{-4}$, a batch size of $128$ and $50$ epochs. The learning rate decreases to $0.2$ of the original every $10$ epochs. We use dlib\footnote{https://github.com/davisking/dlib} as the face detector and expand the region by $1.2$ times to include more facial information.
The temperature $\tau$ in Gumbel Softmax~\cite{jangcategorical} is set to $1$.
For the Semi-supervised \textit{k}-means~\cite{vaze2022generalized} used in Stage 3, $10$ clusters are initialized using K-Means++~\cite{arthur2006k}, with the tolerance of $1e^{-4}$ and max iteration times of $100$.

\textbf{Evaluation Metrics.}
Following \cite{cao2022openworld, han2020automatically, girish2021towards}, we use three metrics to evaluate the performance of all methods on the OW-DFA task, \ie Accuracy (ACC), Normalized Mutual Information (NMI), and Adjusted Rand index (ARI). We align the predicted labels with ground-truth labels using the Hungarian algorithm~\cite{kuhn1955hungarian}. Unless specified, the results we report are obtained through the CPL framework only.

\subsection{Benchmark Evaluation}

\begin{table*}[t]
\caption{Benchmark Evaluation on \textbf{Protocol-1} and \textbf{Protocol-2}.}
\label{tab:compare}
\begin{center}
\resizebox{\linewidth}{!}{%
\begin{tabular}{lcccccccccccccc}
\toprule
\multicolumn{1}{c}{\multirow{3}{*}{\textbf{Method}}} & \multicolumn{7}{c}{\textbf{Protocol-1: Fake}} & \multicolumn{7}{c}{\textbf{Protocol-2: Real \& Fake}} \\ \cmidrule(lr){2-8} \cmidrule(lr){9-15}
\multicolumn{1}{c}{} & \textbf{Known} & \multicolumn{3}{c}{\textbf{Novel}} & \multicolumn{3}{c}{\textbf{All}} & \textbf{Known} & \multicolumn{3}{c}{\textbf{Novel}} & \multicolumn{3}{c}{\textbf{All}} \\ \cmidrule(lr){2-2} \cmidrule(lr){3-5} \cmidrule(lr){6-8}  \cmidrule(lr){9-9} \cmidrule(lr){10-12} \cmidrule(lr){13-15} 
\multicolumn{1}{c}{} & \textbf{ACC} & \textbf{ACC} & \textbf{NMI} & \textbf{ARI} & \textbf{ACC} & \textbf{NMI} & \textbf{ARI} & \textbf{ACC} & \textbf{ACC} & \textbf{NMI} & \textbf{ARI} & \textbf{ACC} & \textbf{NMI} & \textbf{ARI} \\ \midrule

Lower Bound & \textbf{99.96} & 40.96 & 46.43 & 24.05 & 46.90 & 63.18 & 36.35 & \textbf{99.80} & 46.48 & 48.44 & 31.49 & 65.73 & 68.91 & 65.75 \\
Upper Bound & 98.21 & 95.36 & 91.57 & 92.14 & 96.68 & 93.94 & 93.59 & 98.57 & 94.15 & 91.93 & 93.11 & 96.83 & 93.80 & 95.05 \\ \hdashline[2pt/4pt]
DNA-Det~\cite{yang2022deepfake}  & 74.47 & 34.82 & 44.22 & 19.35 & 34.99 & 55.55 & 24.89  & 89.13  & 28.44 & 25.97 & 8.18 & 54.37 & 50.10 & 31.45     \\  
Openworld-GAN~\cite{girish2021towards}  & 99.57 & 38.93 & 45.89 & 41.52 & 57.62 & 57.63 & 47.47 & 99.60 & 46.68 & 53.66 & 45.82 & 69.26 & 58.60 & 61.09 \\ \hdashline[2pt/4pt]
RankStats~\cite{han2020automatically} & 98.58 & 49.94 & 56.05 & 39.76 & 72.49 & 73.63 & 66.49 & 96.84 & 45.26 & 52.44 & 30.17 & 74.39 & 72.21 & 81.66 \\
ORCA~\cite{cao2022openworld} & 97.17 & 66.32 & 63.00 & 53.30 & 80.81 & 79.23 & 74.05 & 95.04 & 53.81 & 60.01 & 38.91 & 78.99 & 78.04 & 83.80 \\
OpenLDN~\cite{rizve2022openldn} & 97.42 & 45.83 & 51.05 & 38.12 & 63.94 & 71.38 & 62.53 & 96.40 & 42.23 & 50.66 & 28.86 & 71.19 & 73.26 & 82.51 \\
NACH~\cite{guo2022robust} & 96.88 & 70.13 & 67.10 & 56.63 & 82.61 & 81.98 & 76.41 & 96.19 & 53.92 & 58.49 & 38.73 & 79.53 & 77.91 & 84.53 \\ \hdashline[2pt/4pt]
\textbf{CPL} & 97.50 & \textbf{71.89} & \textbf{68.20} & \textbf{59.37} & \textbf{83.70} & \textbf{82.31} & \textbf{77.64} & 95.64 & \textbf{59.92} & \textbf{63.90} & \textbf{43.75} & \textbf{81.10} & \textbf{80.23} & \textbf{84.99} \\ \bottomrule
\end{tabular}%
}
\end{center}
\end{table*}

\textbf{Compared Methods.} 
We provide baselines for the OW-DFA task by modifying previous works on GAN attribution~\cite{girish2021towards,yang2022deepfake} and Open-World Semi-Supervised Learning (OW-SSL)~\cite{han2020automatically,cao2022openworld,rizve2022openldn}. We also include the newly released method NACH~\cite{guo2022robust} in our evaluation. To ensure a fair comparison, we use ResNet-50~\cite{he2016deep} as the feature extractor and apply consistent hyperparameters across all approaches. We exclude strong and weak augmentation strategies due to their inapplicability to the OW-DFA task. Additionally, we provide a lower bound based on supervised learning on the labeled set, and an upper bound based on supervised learning on the overall data from both labeled and unlabeled sets. More details are provided in
Appendix Sec.~\ref{sec:cmp}.

\textbf{Results on Protocol-1.} 
We present the results of Protocol-1 in Table~\ref{tab:compare}, demonstrating that CPL outperforms all GAN attribution methods and OW-SSL methods on both novel and overall classes. These results highlight the effectiveness of CPL, surpassing the previous state-of-the-art method NACH~\cite{guo2022robust} by approximately $1.10$-$2.74\%$ absolute improvement on different evaluations for novel classes and $1.09\%$ improvement on ACC for overall classes.
The lower bound experiment, trained only on labeled data, achieves extremely high accuracy for known categories but exhibits poor generalization. Despite the impact of learning novel attacks on prediction results for known attacks, the prediction accuracy of CPL for the known classes remains higher than most OW-SSL methods and only slightly lower than RankStats~\cite{han2020automatically}. However, there is a significant performance gap between RankStats~\cite{han2020automatically} and CPL on the novel and the overall classes.
It is worth noting that DNA-Det~\cite{yang2022deepfake}, a closed-set approach, does not perform well across all classes, as the GAN fingerprints it assumes may not be present in forgery images generated by non-GAN methods. Open-world GAN~\cite{girish2021towards} exceeds the lower bound but does not benefit from semi-supervised learning, limiting the further improvement of its results.

\textbf{Results on Protocol-2.} 
We conduct further experiments on Protocol-2, which incorporates real faces, making the attribution task more challenging and closer to real-world scenarios. Our observations on Protocol-2 are similar to those on Protocol-1, with CPL showing a more significant improvement in the performance of attributing novel and all classes. As shown in Table~\ref{tab:compare}, CPL significantly outperforms NACH~\cite{guo2022robust} and ORCA~\cite{cao2022openworld} by approximately $5.02$-$6.00\%$ and $3.89$-$6.11\%$ respectively on different evaluations of novel classes, while achieving an absolute improvement of $1.57\%$ and $2.11\%$ on ACC for overall classes. Our experiments on Protocol-2 demonstrate that CPL is successful in attributing forged attacks in realistic scenarios containing real data, and is more adept at exploring unlabeled data than existing approaches.

\begin{table}[t]
\caption{Ablation study on each component of CPL on \textbf{Protocol-1}. Each component of CPL contributes towards final performance.}
\vspace{-0.3cm}
\label{tab:ablation_all}
\begin{center}
\resizebox{\columnwidth}{!}{%
\begin{tabular}{ccccccccccc}
\toprule
\multicolumn{1}{c}{\multirow{2}{*}{\textbf{CE}}} &
  \multicolumn{1}{c}{\multirow{2}{*}{\textbf{GR}}} &
  \multicolumn{1}{c}{\multirow{2}{*}{\textbf{GLV}}} &
  \multicolumn{1}{c}{\multirow{2}{*}{\textbf{CSP}}} &
  \multicolumn{1}{c}{\textbf{Known}} &
  \multicolumn{3}{c}{\textbf{Novel}}  &
  \multicolumn{3}{c}{\textbf{All}} \\ \cmidrule(lr){5-5} \cmidrule(lr){6-8} \cmidrule(l){9-11} 
\multicolumn{1}{c}{} &
  \multicolumn{1}{c}{} &
  \multicolumn{1}{c}{} &
  \multicolumn{1}{c}{} &
  \textbf{ACC} &
  \textbf{ACC} &
  \textbf{NMI} &
  \textbf{ARI} &
  \textbf{ACC} &
  \textbf{NMI} &
  \textbf{ARI} \\ \midrule
  
\checkmark &                           &                           &                           & \textbf{99.96} & 40.96 & 46.43 & 24.05 & 46.90 & 63.18 & 36.35 \\
\checkmark & \checkmark &                           &                           & 97.17 & 66.32 & 63.00 & 53.30 & 80.81 & 79.23 & 74.05 \\
\checkmark &  & \checkmark &                           & 96.64 & 68.16 & 66.16 & 57.33 & 81.54 & 81.39 & 77.60 \\ \hdashline[2pt/4pt]
\checkmark & \checkmark &                           & \checkmark & 96.29 & 69.08 & 67.72 & 55.66 & 81.81 & 82.09 & 75.06 \\
\checkmark &  & \checkmark & \checkmark & 97.50 & \textbf{71.89} & \textbf{68.20} & \textbf{59.37} & \textbf{83.70} & \textbf{82.31} & \textbf{77.64} \\ \bottomrule
\end{tabular}%
}
\end{center}
\vspace{-0.4cm}
\end{table}

\subsection{Ablation Study}
\textbf{Components of CPL.} Our analysis of the different components in CPL is presented in Table~\ref{tab:ablation_all}. We first evaluate the performance with cross-entropy loss only, which is consistent with the lower bound experiment. Next, we assess the impact of GLV loss by comparing it to GR loss. Results show that GLV achieves a significant improvement on novel classes, making it the most critical component of our proposed framework. Although GLV loss sacrifices known class performance, we still achieve a substantial enhancement on overall classes. Building on pairwise similarity learning, we further explore the effect of CSP. Comparing the second and fourth rows, we observe that CSP brings an improvement on NMI of $4.72\%$ and $2.86\%$ for novel and overall classes respectively, indicating that CSP can enhance overall performance. Finally, we combine GLV and CSP to obtain the CPL framework, achieving optimal results on both novel and overall classes.
In conclusion, this extensive ablation study empirically validates the effectiveness of different components in CPL.

\begin{figure}[t]
    \centering
    \subfigure[Known-known pairs]{
    \includegraphics[width=0.46\columnwidth]{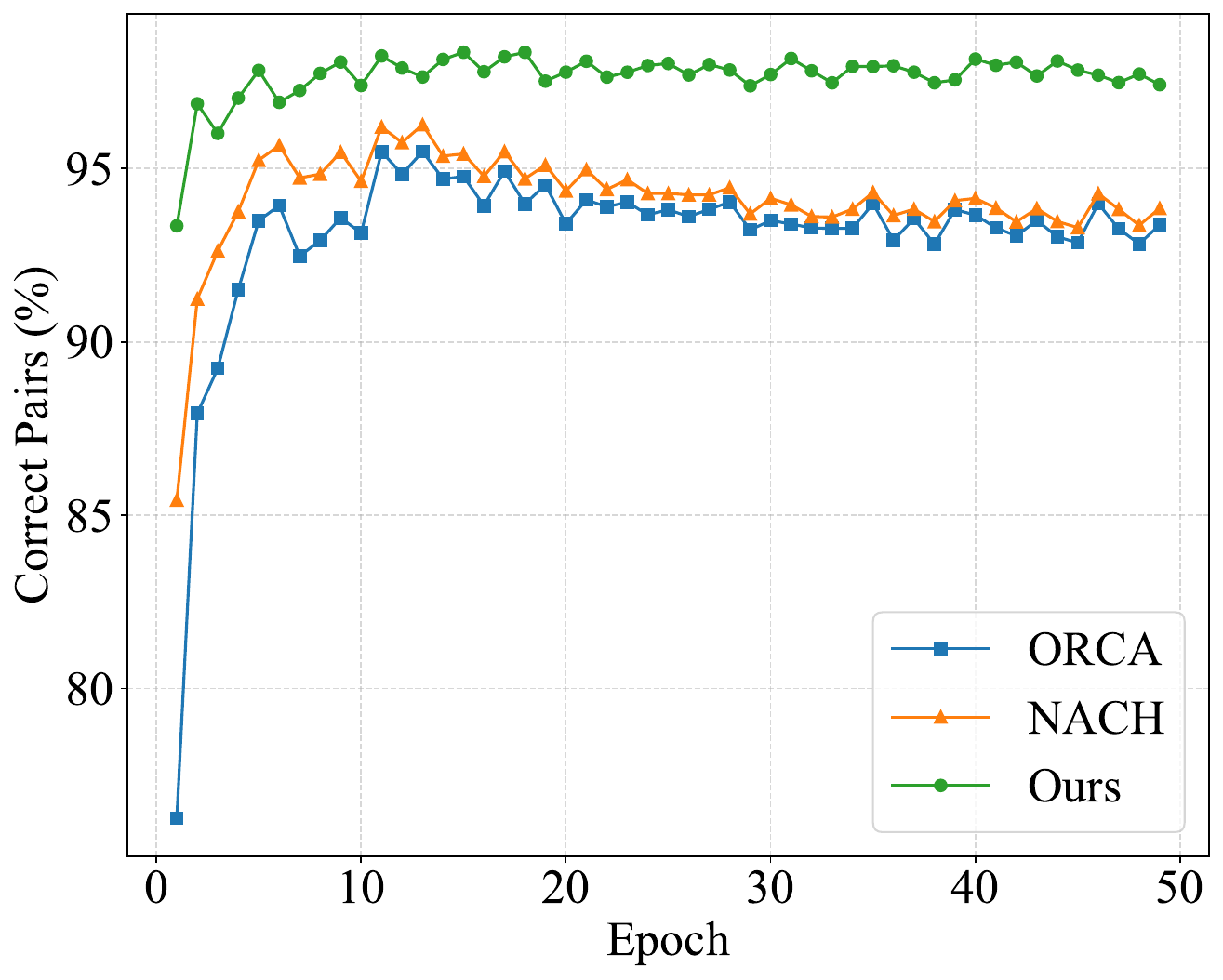}
    }
    \subfigure[Novel-novel pairs]{
    \includegraphics[width=0.46\columnwidth]{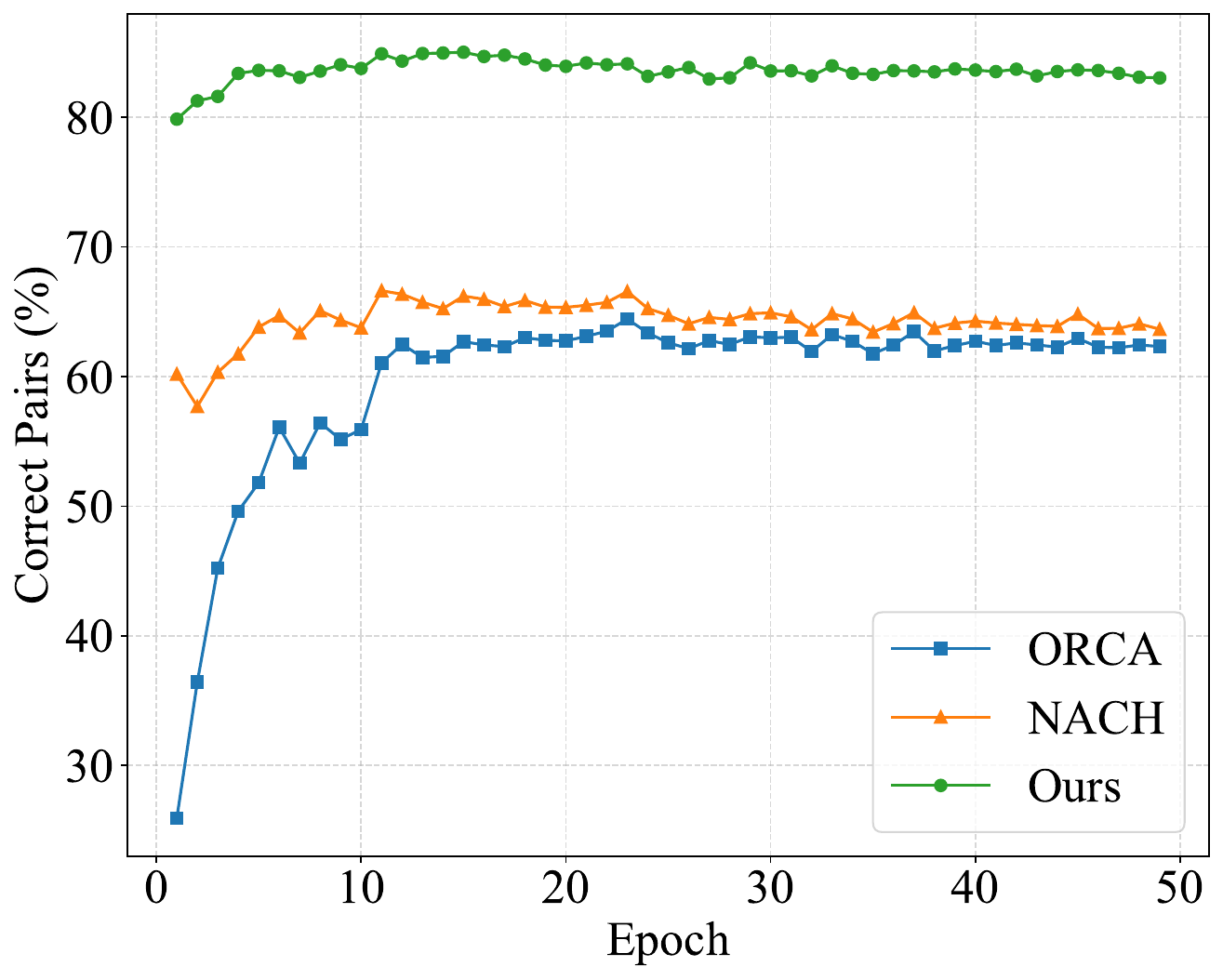}
    }
    \caption{The ratio of correctly selected pairs for known-known pairs and novel-novel pairs with different approaches.}
    \label{fig:abla_rl}
\end{figure}

\textbf{Ablation on GLVM.}
We analyze the correct rate of pairs obtained by different similarity measurement methods for known classes and unknown classes respectively, as shown in Figure~\ref{fig:abla_rl}. From the results, we can see that the GLV loss proposed by us can significantly improve the correct rate of sample matching, both for the known class and for the unknown class. Even in the early stage of training, we can still achieve a correct rate of $\sim\!80\%$ in the accuracy of unknown class matching. Although our recovery rate becomes lower due to the filtering of GLV, it has better results for the overall training by introducing fewer noise samples.

\textbf{Ablation on PPLM.}
We replace the PPLM in the CPL framework with several pseudo-labeling techniques and show the results in Table~\ref{tab:ablation_pl}. We notice that directly assigning labels~\cite{lee2013pseudo} will introduce noisy samples, resulting in a significant decrease in the overall effect. Dynamic-threshold approaches~\cite{zhang2021flexmatch,wang2022freematch} have certain improvements on known classes, but they tend to ignore samples with high uncertainty, causing low performances on novel classes, and fix-threshold approach~\cite{sohn2020fixmatch} also achieves limited improvement. Meanwhile, ST Gumbel Softmax~\cite{jangcategorical} ignores the uncertainty of low-confidence samples. In contrast, our CSP takes all prediction results into account, while reducing the effect of noise by introducing confidence-based weights, achieving the best results on both novel and overall class.

\begin{table}[t]
\centering
\caption{Ablation study on pseudo label strategy on \textbf{Protocol-1}.}
\vspace{-0.6cm}
\label{tab:ablation_pl}
\begin{center}
\resizebox{\columnwidth}{!}{%
\begin{tabular}{lccccccc}
\toprule
\multicolumn{1}{c}{\multirow{2}{*}{\textbf{Pseudo-label Strategy}}} &
  \multicolumn{1}{c}{\textbf{Known}} &
  \multicolumn{3}{c}{\textbf{Novel}} &
  \multicolumn{3}{c}{\textbf{All}} \\ \cmidrule(lr){2-2} \cmidrule(lr){3-5} \cmidrule(l){6-8} 
 &
  \multicolumn{1}{c}{\textbf{ACC}} &
  \multicolumn{1}{c}{\textbf{ACC}} &
  \multicolumn{1}{c}{\textbf{NMI}} &
  \multicolumn{1}{c}{\textbf{ARI}} &
  \multicolumn{1}{c}{\textbf{ACC}} &
  \multicolumn{1}{c}{\textbf{NMI}} &
  \multicolumn{1}{c}{\textbf{ARI}} \\ \midrule
GLV & 96.64 & 68.16 & 66.16 & 57.33 & 81.54 & 81.39 & 77.60 \\ \hdashline[2pt/4pt]
GLV + Pseudo-label~\cite{lee2013pseudo} & 96.61 & 65.55 & 65.18 & 55.66 & 80.14 & 80.78 & 76.78 \\
GLV + FixMatch~\cite{sohn2020fixmatch} & 96.14 & 67.21 & 66.18 & 56.33 & 80.81 & 80.82 & 76.33 \\
GLV + FlexMatch~\cite{zhang2021flexmatch} & 96.64 & 66.16 & 65.57 & 56.10 & 80.48 & 81.06 & 76.99 \\
GLV + FreeMatch~\cite{wang2022freematch} & 97.02 & 67.74 & 67.00 & 56.45 & 81.50 & 81.43 & 76.85 \\
GLV + Gumbel-Softmax~\cite{jangcategorical} & 96.33 & 68.27 & 67.92 & 57.69 & 81.46 & 82.11 & 77.49 \\ \hdashline[2pt/4pt]
\textbf{GLV + CSP (CPL)} & \textbf{97.50} & \textbf{71.89} & \textbf{68.20} & \textbf{59.37} & \textbf{83.70} & \textbf{82.31} & \textbf{77.64}\\ \bottomrule
\end{tabular}%
}
\end{center}
\end{table}

\begin{table}[t]
\centering
\caption{Results of multi-stage paradigm on \textbf{Protocol-1}.}
\vspace{0.1cm}
\label{tab:stage}
\resizebox{\columnwidth}{!}{%
\begin{tabular}{lccccccc}
\toprule
\multicolumn{1}{c}{\multirow{2}{*}{\textbf{Stage}}} & \textbf{Known} & \multicolumn{3}{c}{\textbf{Novel}} & \multicolumn{3}{c}{\textbf{All}} \\ \cmidrule(lr){2-2}  \cmidrule(lr){3-5} \cmidrule(lr){6-8}
\multicolumn{1}{c}{} & \textbf{ACC} & \textbf{ACC} & \textbf{NMI} & \textbf{ARI} & \textbf{ACC} & \textbf{NMI} & \textbf{ARI} \\ \midrule
S1-Pretrain & \textbf{99.96} & 40.96 & 46.43 & 24.05 & 46.90 & 63.18 & 36.35 \\
S2-CPL & 97.33 & 71.75 & 67.68 & 58.03 & 83.59 & 82.36 & 77.47 \\
S3-IL & 97.08 & \textbf{72.78} & \textbf{70.87} & \textbf{59.10} & \textbf{84.20} & \textbf{84.05} & \textbf{77.58} \\ \bottomrule
\end{tabular}%
}
\end{table}

\textbf{Multi-stage Paradigm.}
We further conduct an ablation study to evaluate the performance of the Multi-stage Paradigm in Algorithm~\ref{alg:framework} and report the results in Table~\ref{tab:stage}. Stage 1 is exactly the lower bound of our method in Table~\ref{tab:compare}. The results of Stage 2 suggest that initializing the model with pre-trained weights on the labeled set accelerates the semi-supervised learning process, while providing a slight improvement in effectiveness compared to direct training based on weights pre-trained on ImageNet~\cite{deng2009imagenet}. For the iterative learning in Stage 3, we use Semi-supervised K-Means~\cite{vaze2022generalized} to generate pseudo labels and apply fine-tuning on the previous model, further improving the performance of the model on both novel and overall classes. Table~\ref{tab:stage} demonstrates that each stage in the paradigm contributes to the high performance of our method.

\begin{figure}[t]
    \centering
    \subfigure[Openworld-GAN\cite{girish2021towards}]{
    \includegraphics[width=0.4\columnwidth]{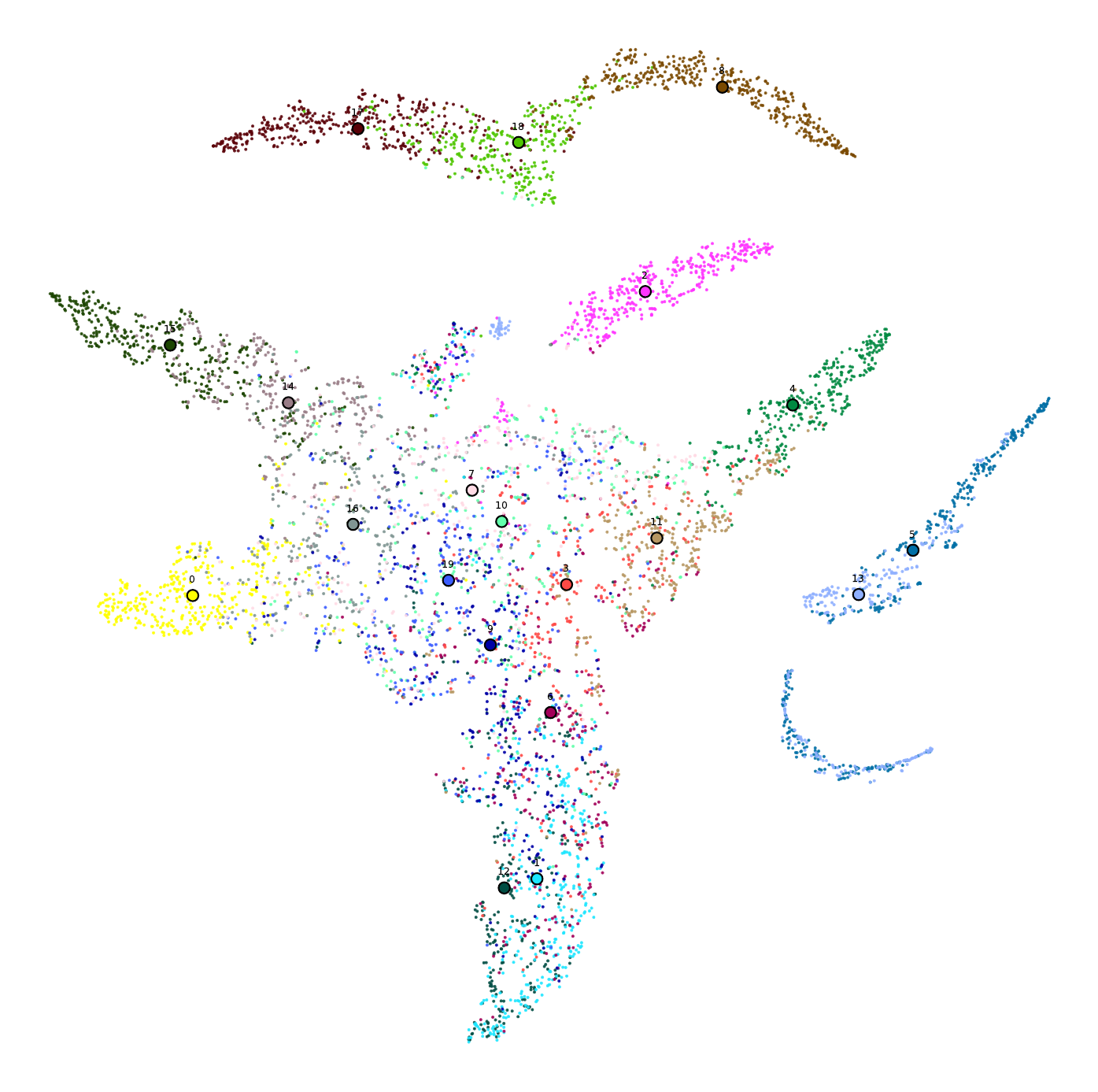}
    }
    \subfigure[CPL]{
    \includegraphics[width=0.52\columnwidth]{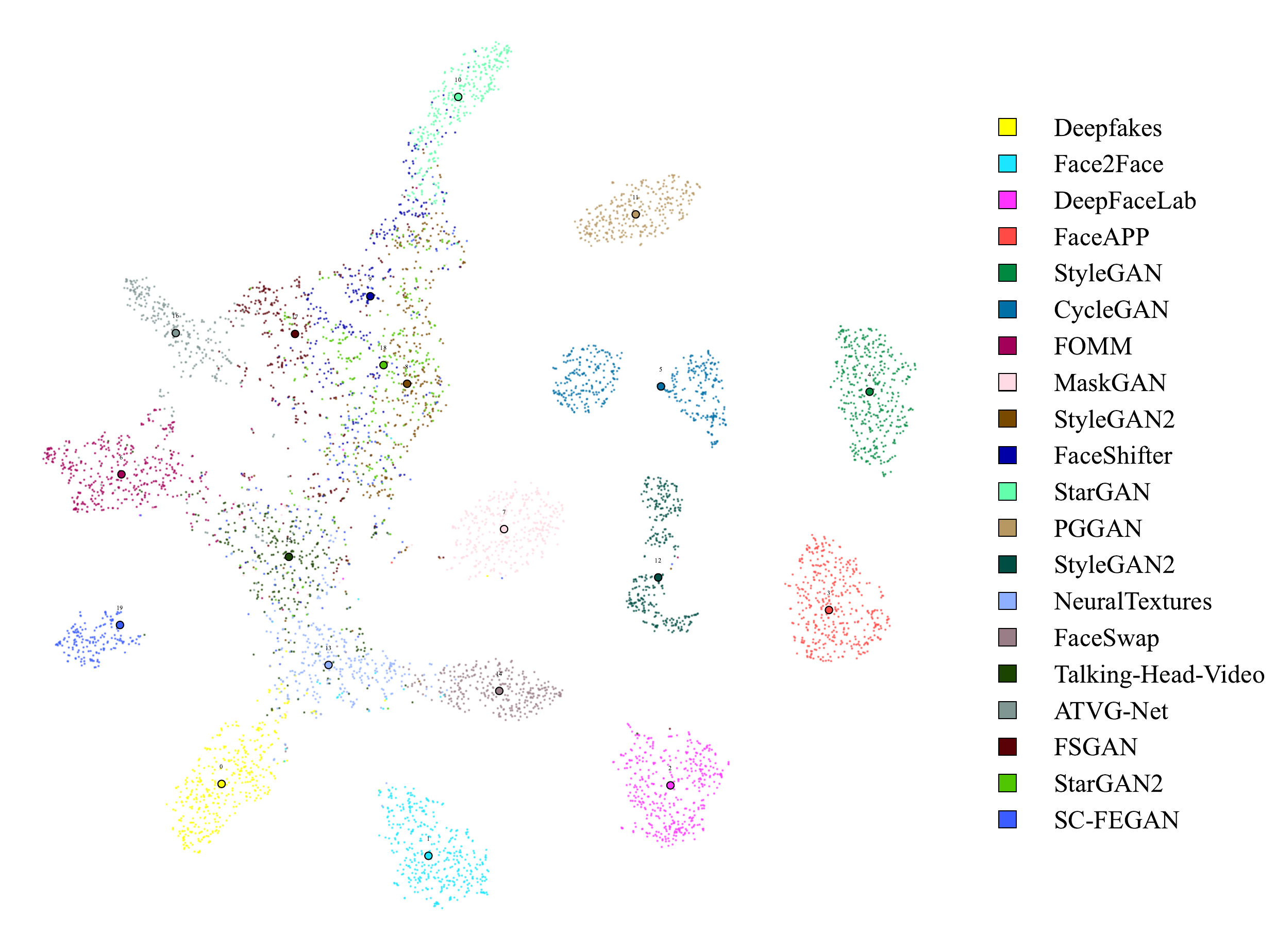}
    }
    \caption{t-SNE visualization on \textbf{Protocol-1}.}
    \label{fig:tsne}
\end{figure}

\textbf{t-SNE Visualization.}
In order to compare the performance of CPL more intuitively, we performed t-SNE~\cite{tsne} visualization for Open-World GAN~\cite{girish2021towards} and CPL in Figure~\ref{fig:tsne}. We observe that CPL has greatly improved the clustering performance compared to Open-World GAN~\cite{girish2021towards}. Given the satisfactory results of known classes, CPL is capable of isolating novel classes of lower difficulty into distinct classes, \textit{e.g.}, StyleGAN2~\cite{Karras2019stylegan2} and SC-FEGAN~\cite{Jo_2019_ICCV}. For novel attacks of higher difficulty, CPL is also effective in clustering these samples. On the other hand, the gap between different attack types is significantly larger, even for data within the same dataset, \textit{e.g.}, Deepfakes~\cite{deepfakes}, FaceSwap~\cite{faceswap} and NeuralTextures~\cite{NeuralTextures} in FF++~\cite{rossler2019ff++}, and this can be attributed to the fact that CPL concentrates on combining patch-wise local similarity with global similarity. 

\begin{table}[t]
\centering
\caption{AUC results for real/fake detection on \textbf{Protocol-2}.}
\vspace{0.2cm}
\label{tab:auc}
\resizebox{\columnwidth}{!}{%
\begin{tabular}{cccccc}
\toprule
\multicolumn{3}{c}{\textbf{Data}} & \multicolumn{3}{c}{\textbf{Approach}} \\ \cmidrule(lr){1-3} \cmidrule(lr){4-6}
\textbf{Known} & \textbf{New Fake} & \textbf{New Real} & \textbf{a) Binary} & \textbf{b) Multi} & \textbf{c) CPL} \\ \midrule
\checkmark &  &  & 99.95 & 100.00 & \textbf{100.00} \\
\checkmark & \checkmark &  & 93.06 & 94.80 & \textbf{99.91} \\
\checkmark & \checkmark & \checkmark & 94.99 & 95.84 & \textbf{96.28} \\
 & \checkmark & \checkmark & 73.91 & 76.47 & \textbf{85.97} \\ \bottomrule
\end{tabular}%
}
\end{table}

\subsection{Real/Fake Detection}
To further verify the significance of the deepfake attribution task for deepfake detection, we conduct additional experiments for comparison based on Protocol-2. We compare the results of three approaches: a) deepfake binary classification, b) deepfake multi-classification, and c) our CPL framework. Approaches a) and b) are trained on labeled data, while the CPL framework utilizes both labeled and unlabeled data.
We construct unlabeled sets using various combinations of known, new fake, and new real faces. The AUC results of these approaches are evaluated and presented in Table~\ref{tab:auc}.
We observe that the performance of b) is consistently higher than that of a), especially when new images are introduced. Compared to b) and c), the CPL framework achieves a significant improvement with $\sim\!9.5\%$ AUC on new fake and new real set. These results clearly illustrate that the introduction of the deepfake attribution task can further enhance the security of the deepfake detection task.




\section{Conclusion}
We introduce a novel benchmark, Open-World DeepFake Attribution (OW-DFA), which aims to enhance attribution performance against various types of fake faces in open-world scenarios. Our proposed framework, Contrastive Pseudo Learning (CPL), introduces a Global-Local Voting module to guide the inter-sample relations of forged faces with different manipulated regions. A probability-based pseudo-label strategy is also employed to mitigate the pseudo-noise caused by similar attack methods. Furthermore, we extend the CPL framework with a multi-stage paradigm that incorporates pre-training techniques and iterative learning to further improve traceability performance. Extensive experiments demonstrate the superiority of CPL on the OW-DFA benchmark. We also highlight the interpretability and security of the DFA task and its impact on the deepfake detection field.

\section*{Acknowledgements}

This project is supported by
National Natural Science Foundation of China (No.72192821,61972157,62272447),
Shanghai Municipal Science and Technology Major Project (2021SHZDZX0102),
Shanghai Science and Technology Commission  (21511101200),
Shanghai Sailing Program (22YF1420300),
CCF-Tencent Open Research Fund (RAGR20220121),
Young Elite Scientists Sponsorship Program by CAST (2022QNRC001),
Beijing Natural Science Foundation (L222117),
the Fundamental Research Funds for the Central Universities (YG2023QNB17)

{\small
\bibliographystyle{ieee_fullname}
\bibliography{egbib}
}

\clearpage
\section*{\LARGE Appendix}
\begin{table*}[ht]
\centering
\caption{Relationship between our novel OW-DFA and related tasks.}
\label{tab:setting_compare}
\resizebox{\textwidth}{!}{%
\begin{tabular}{lcccc}
\toprule
\multicolumn{1}{c}{\textbf{Task}} & \multicolumn{1}{c}{\textbf{Task Goal}} & \multicolumn{1}{c}{\textbf{Data Type}} & \textbf{Known Classes} & \textbf{Novel Classes} \\ \midrule
Deepfake Detection & Classification of real/fake faces & Deepfake & \checkmark & - \\
GAN Attribution & Classification of GAN images & GAN-generated & \checkmark & - \\
Open-world GAN Attribution & Classification of GAN images & GAN-generated & \checkmark & \checkmark \\
Open-world Semi-Supervised Learning & Classification of object & Various object images & \checkmark & \checkmark \\
Open-world DeepFake Attribution & Classification of deepfake faces & Deepfake & \checkmark & \checkmark \\ \bottomrule
\end{tabular}%
}
\end{table*}

\appendix

\vspace{0.1cm}

\section{Comparison of OW-DFA with other Tasks}

We summarize similarities and differences between OW-DFA and related tasks in Table~\ref{tab:setting_compare}.

\textbf{Comparison with GAN attribution.}
GAN attribution~\cite{yu2019attributing, yu2021artificial, yang2022deepfake, guarnera2022exploitation} is a multi-classification task focusing on attributing GAN models. A common strategy is to use the fingerprints of different GAN models to attribute those generated images. However, they only consider the close-world scenario where the training and test sets have the same category distribution. Such an assumption is not applicable in OW-DFA, since novel forgeries emerge greatly under open-world scenarios.

\textbf{Comparison with OW-GAN attribution.}
OW-GAN attribution is a multi-classification task that focuses on attributing GAN models and discovering unseen GANs in an open-world scenario, which is proposed by Open-world GAN~\cite{girish2021towards}. Although some progress has been made in open-world scenarios, the fingerprint assumption it relied on may not hold in the fake faces generated by non-GAN methods. Besides GAN methods, OW-DFA also covers other forgery types, including \textit{identity swap} and \textit{expression transfer}, making the task more realistic and challenging.

\textbf{Comparison with Deepfake Detection.} 
Deepfake detection focuses on real/fake detection, and many related works~\cite{qian2020thinking,zhao2021multi} have been proposed in recent years. However, the generalization performance on novel attacks is still limited. As fake faces become visually realistic and need to be interpreted in legal proceedings, OW-DFA extends the binary detection task to a multi-classification task for enhancing the interpretability of deepfake detection. At the same time, the additional provision of unlabeled novel attack data also provides a higher possibility for further improvement of generalizability.

\begin{table}[h]
\centering
\caption{List of forgery methods and corresponding train/test splits used in \textbf{Protocol-1} and \textbf{Protocol-2}. Note that some train images are unlabeled.}
\vspace{1mm}
\label{tab:split}
\resizebox{\columnwidth}{!}{%
\begin{tabular}{ccccc}
\toprule
\textbf{Face Type} & \textbf{Source Dataset} & \textbf{Method} & \multicolumn{1}{l}{\textbf{\# of Train}} & \multicolumn{1}{l}{\textbf{\# of Test}} \\ \midrule
\multirow{5}{*}{Identity Swap} & \multirow{2}{*}{FaceForensics++} & FaceSwap & 1200 & 300 \\
 &  & Deepfakes & 1600 & 400 \\ \cmidrule(l){2-5} 
 & \multirow{3}{*}{ForgeryNet} & FaceShifter & 1200 & 300 \\
 &  & DeepFaceLab & 1600 & 400 \\
 &  & FSGAN & 1200 & 300 \\ \midrule
\multirow{5}{*}{Expression Transfer} & \multirow{2}{*}{FaceForensics++} & Face2Face & 1600 & 400 \\
 &  & NeuralTextures & 1200 & 300 \\ \cmidrule(l){2-5} 
 & \multirow{3}{*}{ForgeryNet} & Talking-Head-Video & 1200 & 300 \\
 &  & ATVG-Net & 1200 & 300 \\
 &  & FOMM & 1600 & 400 \\ \midrule
\multirow{5}{*}{Attribute Manipulation} & \multirow{3}{*}{ForgeryNet} & MaskGAN & 1600 & 400 \\
 &  & StarGAN2 & 1200 & 300 \\
 &  & SC-FEGAN & 1200 & 300 \\ \cmidrule(l){2-5} 
 & \multirow{2}{*}{DFFD} & FaceAPP & 1600 & 400 \\
 &  & StarGAN & 1200 & 300 \\ \midrule
\multirow{5}{*}{Entire Face Syncthesis} & ForgeryNet & StyleGAN2 & 1200 & 300 \\ \cmidrule(l){2-5} 
 & \multirow{2}{*}{DFFD} & PGGAN & 1200 & 300 \\
 &  & StyleGAN & 1600 & 400 \\ \cmidrule(l){2-5} 
 & \multirow{2}{*}{ForgeryNIR} & CycleGAN & 1600 & 400 \\
 &  & StyleGAN2 & 1200 & 300 \\ \midrule
\multirow{2}{*}{Real Face} & FaceForensics++ & Youtube-Real & 16000 & 4000 \\ \cmidrule(l){2-5} 
 & CelebDFv2 & Celeb-Real & 4000 & 1000 \\ \bottomrule
\end{tabular}%
}
\end{table}

\vspace{0.1cm}

\section{Pre-processing Details of Datasets\label{sec:prepro}}

We present the five datasets that are used in our OW-DFA benchmark and describe the detail of data processing for each dataset.

\begin{itemize}
    \item \textbf{FaceForensics++}~\cite{rossler2019ff++} is the most widely used dataset for deepfake detection tasks, consisting of $1,\!000$ original video sequences that have been manipulated with $4$ face manipulation methods, including Deepfakes, Face2Face, FaceSwap, and NeuralTextures. As part of the data for OW-DFA, we include both real and fake images from FF++. We sample $20$ frames for each manipulated video and $200$ frames for each original video. After that, we use dlib to crop out the faces from those frames and save them as new images.
    \item \textbf{CelebDF}~\cite{li2020celeb} is a challenging dataset for deepfake detection. It consists of $590$ celebrity videos (Celeb-real) and $300$ additional videos (YouTube-real) downloaded from YouTube, as well as $5,\!639$ high-quality synthesized videos. The inclusion of real celebrity videos in CelebDF makes it suitable for evaluating the OW-DFA benchmark under Protocol-2, which requires distinguishing between real and fake images from different sources. We sample $100$ frames for each Celeb-real video and use dlib to crop the faces at the same time.
    \item \textbf{ForgeryNet}~\cite{he2021forgerynet} is the largest publicly available multi-purpose deep face forgery analysis benchmark dataset. It contains $2.9$ million images and $15$ forgery methods. Due to its large scale and diverse range of attack types, ForgeryNet is the most suitable dataset for deepfake attribution tasks. A significant portion of the data in the OW-DFA benchmark is obtained from ForgeryNet. For each forgery method in Protocol-1, we extract $20,\!000$ frames and apply dlib to ensure data consistency.
    \item \textbf{DFFD}~\cite{dang2020detection} is a diverse deepfake face dataset that contains $600,\!000$ face images. Of these images, $500,\!000$ are synthetic or manipulated and $100,\!000$ are real. The images originate from various publicly accessible datasets and are synthesized or manipulated using publicly accessible methods. Owing to its diversity of attack types and inclusion of data on attribute manipulation and entire face synthesis, DFFD is incorporated into the OW-DFA benchmark. For FaceAPP and GAN generation attacks, we randomly select $20,\!000$ images for each method.
    \item \textbf{ForgeryNIR}~\cite{wang2022forgerynir} is a near-infrared face forgery and detection dataset that contains over $50,\!000$ real and fake identities. It also includes various perturbations to simulate real-world scenarios. Since the fake images in ForgeryNIR are generated using multiple GAN techniques, we randomly select $20,\!000$ images for both CycleGAN and StyleGAN2 and include them in OW-DFA.
\end{itemize}

\textbf{Train and test splits.}
We download all datasets from the official links. We select images according to \textbf{Protocol-1} ($20$ manipulation methods) and \textbf{Protocol-2} ($20$ manipulation methods and $2$ real face types). Then, we randomly sample images according to the corresponding number of each forgery attack method. Train and test sets are split based on the ratio of $4\!:\!1$.
Table~\ref{tab:split} summarizes the class-wise train and test splits used in Protocol-1 and Protocol-2. Note that some train images are unlabeled. 
Protocol-1 covers $20$ forgery methods and includes a total of $272,\!000$ training images and $68,\!000$ test images.
Protocol-2 covers both $2$ real face and $20$ forgery methods and includes a total of $472,\!000$ training images and $118,\!000$ test images.

\section{Implementation for Multi-stage Paradigm\label{sec:msp}}
To further improve the performance of the OW-DFA task, we extend CPL to a multi-stage paradigm with a pretraining technique and iterative learning. Here we also provide the specific implementation details of different stages.

\begin{itemize}

    \item \textbf{Stage-1} aims to pre-train on the labeled dataset to improve the performance on known attacks. Specifically, we conduct supervised training based on the labeled data in OW-DFA using Eq.~11 in the main text as the loss function with a learning rate of $2e^{-4}$ for $20$ epochs. After completing Stage-1 training, we obtain a weight that performs well on known attacks and can be used as the pretrained weight for Stage-2.
    
    \item \textbf{Stage-2} aims to leverage the unlabeled data to enhance the robustness and generalization of the model. We initialize the model with the pretrained weights from Stage-1 and apply CPL on both labeled and unlabeled sets in a semi-supervised manner. We use Eq.~13 in the main text as the loss function with a learning rate of $2e^{-4}$ for $50$ epochs. We also ensure a strict half-sampling of labeled and unlabeled data in a batch, maintaining a balanced ratio of the two types of data during training.

    \item \textbf{Stage-3} aims to exploit the clustering structure of the feature space and assign more accurate labels to the unlabeled data. We leverage the Semi-Supervised \textit{k}-means algorithm~\cite{vaze2022generalized} and iterative learning to further refine the pseudo-labels and fine-tune the model. We first extract features of all training samples, both labeled and unlabeled, with the feature extractor in Stage-2. Next, we set up initial clustering centers with $10$ samples by K-Means++. Then, Semi-Supervised \textit{k}-means (refer to Figure~4 in~\cite{vaze2022generalized}) will be repeated for at most $100$ iteration times until the k-means algorithm converges with a tolerance of $1e^{-4}$. After obtaining pseudo-labels with assigned clusters, we further fine-tune our models using Eq.~11 in the main text as the loss function with a learning rate of $2e^{-4}$ for $20$ epochs.
    
\end{itemize}

\section{Implementation for Comparison Methods\label{sec:cmp}}
Our baseline comparison includes a total of $8$ methods, comprising GAN attribution and OW-SSL methods. To ensure a fair comparison between methods, we use the actual number of categories as the output head number for the classifier. We exclude strong and weak augmentation strategies due to their inapplicability to the OW-DFA task. All methods use ResNet50 pre-trained on ImageNet as the feature extractor. It is trained with a learning rate of $2e^{-4}$ for $50$ epochs and a batch size of $128$.

\begin{itemize}
    \item \textbf{Lower bound} is established using supervised learning on labeled data. Since this experiment applies to a closed-world setting, we obtain the output result based on its original classifier and evaluate its performance directly.
    \item \textbf{Upper bound} is established using supervised learning on all data, including both labeled and unlabeled data. Since this experiment is trained with all types of samples exposed, its performance must be optimal.
    \item \textbf{DNA-Det}~\cite{yang2022deepfake} is a closed-set approach that attributes GAN-generated images based on GAN fingerprints. We include classification loss, contrastive loss, and automatic weighted loss with default configurations.
    \item \textbf{Open-World GAN}~\cite{girish2021towards} is an approach that discovers and attributes GAN-generated images based on an open-world setting. We config the class lists of both protocols and repeat iteration for $4$ times according to the default configuration. We extend evaluation to an additional test set and report results on this extra set.
    \item \textbf{RankStats}~\cite{han2020automatically} is a novel class discovery method that can be extended to solve OW-SSL tasks by exploring Top-K ranked dimensions of features. Sample pairs can be pulled or pushed based on their similarity. We use the default setting of $K\!=\!5$ as the number of ranked dimensions.
    \item \textbf{ORCA}~\cite{cao2022openworld} is the first approach to propose the task of OW-SSL and uses cosine distance as a similarity matrix to bring pairs with high similarity closer together. We reproduce ORCA with both a fixed negative margin and a dynamic margin and report the best result with a fixed negative margin of $m=-0.2$.
    \item \textbf{OpenLDN}~\cite{rizve2022openldn} uses a bi-level optimization rule to enhance feature representation and applies close-world iterative training to improve performance. However, we only evaluate its performance using its semi-supervised feature learning component. We change the backbone to ResNet50 while keeping the configuration of simnet unchanged, and use 0.5 as the default threshold for pseudo-label assignment.
    \item \textbf{NACH}~\cite{guo2022robust} is a recently introduced approach that improves ORCA’s performance by filtering out erroneous samples and synchronizing the learning pace between seen and unseen classes. We use the default setting of $K=2$ as the index for the labeled sample when filtering pairs.
\end{itemize}

\begin{table*}[th]
\begin{center}
\caption{List of methods and corresponding datasets utilized in OW-DFA with $5 \times$ scale.}
\vspace{-0.1cm}
\label{tab:dataset_5x}
\resizebox{\textwidth}{!}{%
\begin{tabular}{@{}cllccccc@{}}
\toprule
\textbf{Face Type} &
  \textbf{Labeled Sets} &
  \textbf{Unlabeled Sets} &
  \textbf{Source Dataset} &
  \textbf{Method} &
  \textbf{Tag} &
  \textbf{Labeled \#} &
  \textbf{Unlabeled \#} \\ \midrule
\multirow{5}{*}[-2pt]{Identity Swap} &
  \multirow{5}{*}[-2pt]{\begin{tabular}[c]{@{}l@{}}Deepfakes~\cite{deepfakes}\\ DeepFaceLab~\cite{DeepFaceLab}\end{tabular}} &
  \multirow{5}{*}[-2pt]{\begin{tabular}[c]{@{}l@{}}Deepfakes\\ DeepFaceLab\\ FaceSwap~\cite{faceswap}\\ FaceShifter~\cite{li2019faceshifter}\\ FSGAN~\cite{nirkin2019fsgan}\end{tabular}} &
  \multirow{2}{*}{FaceForensics++~\cite{rossler2019ff++}} &
  Deepfakes & Known & 7500 & 2500 \\
 &  &  & & FaceSwap & Novel & - & 7500 \\ \cmidrule(lr){4-8}
 &  &  & \multirow{3}{*}{ForgeryNet~\cite{he2021forgerynet}} & DeepFaceLab & Known & 7500 & 2500 \\
 &  &  & & FaceShifter & Novel & - & 7500 \\
 &  &  & & FSGAN & Novel & - & 7500\\ \midrule
\multirow{5}{*}[-2pt]{Expression Transfer} &
  \multirow{5}{*}[-2pt]{\begin{tabular}[c]{@{}l@{}}Face2Face~\cite{face2face}\\ FOMM~\cite{fomm}\end{tabular}} &
  \multirow{5}{*}[-2pt]{\begin{tabular}[c]{@{}l@{}}Face2Face\\ FOMM \\ NeuralTextures~\cite{NeuralTextures}\\ Talking-Head-Video~\cite{zhang2021text2video}\\ ATVG-Net~\cite{chen2019hierarchical}\end{tabular}} &
  \multirow{2}{*}{FaceForensics++} & Face2Face & Known & 7500 & 2500  \\
 &  &  & & NeuralTextures     & Novel & -    & 7500 \\ \cmidrule(lr){4-8}
 &  &  & \multirow{3}{*}{ForgeryNet} & FOMM               & Known & 7500& 2500   \\
 &  &  & & ATVG-Net           & Novel & -    & 7500 \\
 &  &  & & Talking-Head-Video & Novel & -    & 7500 \\ \midrule
\multirow{5}{*}[-2pt]{Attribute Manipulation} &
  \multirow{5}{*}[-2pt]{\begin{tabular}[c]{@{}l@{}}MaskGAN~\cite{CelebAMaskGAN-HQ}\\ FaceAPP~\cite{faceapp}\end{tabular}} &
  \multirow{5}{*}[-2pt]{\begin{tabular}[c]{@{}l@{}}MaskGAN\\ FaceAPP\\ StarGAN2~\cite{choi2020starganv2}\\ SC-FEGAN~\cite{Jo_2019_ICCV}\\ StarGAN~\cite{choi2018stargan}\end{tabular}} &
  \multirow{3}{*}{ForgeryNet} & MaskGAN & Known & 7500 & 2500 \\
 &  &  & & StarGAN2 & Novel & - & 7500 \\
 &  &  & & SC-FEGAN & Novel & - & 7500 \\ \cmidrule(lr){4-8}
 &  &  & \multirow{2}{*}{DFFD~\cite{dang2020detection}} & FaceAPP & Known & 7500 & 2500 \\
 &  &  & & StarGAN & Novel & - & 7500 \\ \midrule
\multirow{5}{*}[-4pt]{Entire Face Synthesis} &
  \multirow{5}{*}[-4pt]{\begin{tabular}[c]{@{}l@{}}StyleGAN~\cite{karras2019style}\\ CycleGAN~\cite{CycleGAN2017}\end{tabular}} &
  \multirow{5}{*}[-4pt]{\begin{tabular}[c]{@{}l@{}}StyleGAN\\ CycleGAN\\ PGGAN~\cite{karras2018progressive}\\ StyleGAN2~\cite{Karras2019stylegan2}\end{tabular}} &
  ForgeryNet & StyleGAN2 & Novel & - & 7500\\ \cmidrule(lr){4-8}
 &  &  & \multirow{2}{*}{DFFD} & StyleGAN & Known & 7500 & 2500  \\
 &  &  & & PGGAN & Novel &  -   & 7500 \\ \cmidrule(lr){4-8}
 &  &  & \multirow{2}{*}{ForgeryNIR~\cite{wang2022forgerynir}} & CycleGAN & Known & 7500 & 2500 \\ 
 &  &  & & StyleGAN2 & Novel &  -   & 7500 \\ \midrule 
\multirow{2}{*}[-2pt]{Real Face} & \multirow{2}{*}[-2pt]{\begin{tabular}[c]{@{}l@{}} Youtube-Real~\cite{rossler2019ff++}\end{tabular}} & \multirow{2}{*}[-2pt]{\begin{tabular}[c]{@{}l@{}} Celeb-Real~\cite{li2020celeb}\end{tabular}} & FaceForensics++ & Youtube-Real & Known & 75000 & 25000 \\ \cmidrule(lr){4-8}
 & & & CelebDFv2~\cite{li2020celeb} & Celeb-Real & Novel & - & 25000 \\ \bottomrule
\end{tabular}%
}
\end{center}
\vspace{-0.2cm}
\end{table*}

\begin{table*}[t]
\caption{Benchmark Evaluation on \textbf{Protocol-1} and \textbf{Protocol-2} with dataset of $5\times$ scale.}
\vspace{0.2cm}
\label{tab:compare5}
\begin{center}
\resizebox{\linewidth}{!}{%
\begin{tabular}{lcccccccccccccc}
\toprule
\multicolumn{1}{c}{\multirow{3}{*}{\textbf{Method}}} & \multicolumn{7}{c}{\textbf{Protocol-1: Fake}} & \multicolumn{7}{c}{\textbf{Protocol-2: Real \& Fake}} \\ \cmidrule(lr){2-8} \cmidrule(lr){9-15}
\multicolumn{1}{c}{} & \textbf{Known} & \multicolumn{3}{c}{\textbf{Novel}} & \multicolumn{3}{c}{\textbf{All}} & \textbf{Known} & \multicolumn{3}{c}{\textbf{Novel}} & \multicolumn{3}{c}{\textbf{All}} \\ \cmidrule(lr){2-2} \cmidrule(lr){3-5} \cmidrule(lr){6-8}  \cmidrule(lr){9-9} \cmidrule(lr){10-12} \cmidrule(lr){13-15} 
\multicolumn{1}{c}{} & \textbf{ACC} & \textbf{ACC} & \textbf{NMI} & \textbf{ARI} & \textbf{ACC} & \textbf{NMI} & \textbf{ARI} & \textbf{ACC} & \textbf{ACC} & \textbf{NMI} & \textbf{ARI} & \textbf{ACC} & \textbf{NMI} & \textbf{ARI} \\ \midrule

Lower Bound & \textbf{99.68} & 40.86 & 47.55 & 26.33 & 46.91 & 63.43 & 37.33 & \textbf{99.84} & 34.57 & 42.98 & 19.37 & 61.46 & 66.05 & 62.16 \\
Upper Bound & 98.93 & 96.99 & 94.18 & 94.94 & 97.91 & 95.87 & 95.91 & 99.27 & 97.12 & 94.89 & 96.78 & 98.43 & 96.48 & 98.27 \\ \hdashline[2pt/4pt]
RankStats~\cite{han2020automatically} & 99.17 & 62.05 & 64.60 & 52.87 & 79.52 & 78.87 & 72.90 & 98.86 & 51.19 & 57.56 & 37.56 & 78.25 & 77.37 & 88.07 \\
ORCA~\cite{cao2022openworld} & 98.30 & 73.61 & 70.20 & 63.50 & 85.23 & 83.99 & 80.86 & 97.09 & 62.10 & 64.96 & 49.15 & 83.44 & 82.68 & 88.64 \\
OpenLDN~\cite{rizve2022openldn}   & 98.78 & 54.12 & 57.54 & 45.43 & 72.90 & 77.22 & 70.03 & 97.03 & 48.26 & 52.77 & 33.72 & 73.97 & 75.13 & 84.37 \\
NACH~\cite{guo2022robust} & 98.34 & 73.43 & 71.61 & 65.33 & 85.16 & 84.90 & 82.31 & 97.28 & 69.39 & 70.03 & 54.28 & 86.47 & 84.76 & 90.09 \\ \hdashline[2pt/4pt]
CPL & 98.68 & \textbf{75.21} & \textbf{73.19} & \textbf{65.71} & \textbf{86.25} & \textbf{85.58} & \textbf{82.35} & 97.45 & \textbf{69.57} & \textbf{70.67} & \textbf{54.67} & \textbf{86.51} & \textbf{85.44} & \textbf{90.30} \\ \bottomrule
\end{tabular}%
}
\end{center}
\vspace{-0.5cm}
\end{table*}

\begin{figure*}[t]
    \centering
    \subfigure[Result on \textbf{Protocol-1}]{\includegraphics[width=0.48\linewidth]{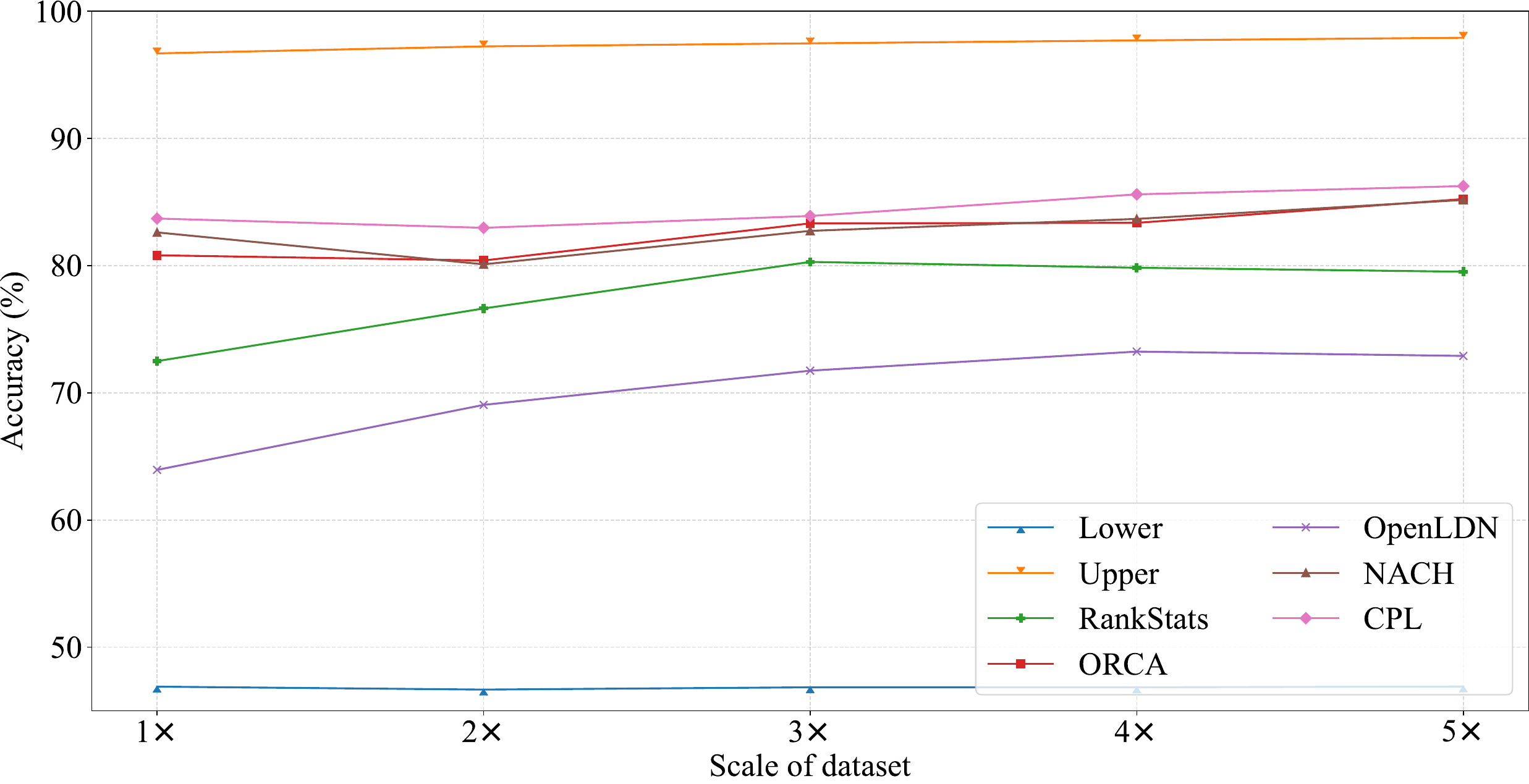}}
    \hfill
    \subfigure[Result on \textbf{Protocol-2}]{\includegraphics[width=0.48\linewidth]{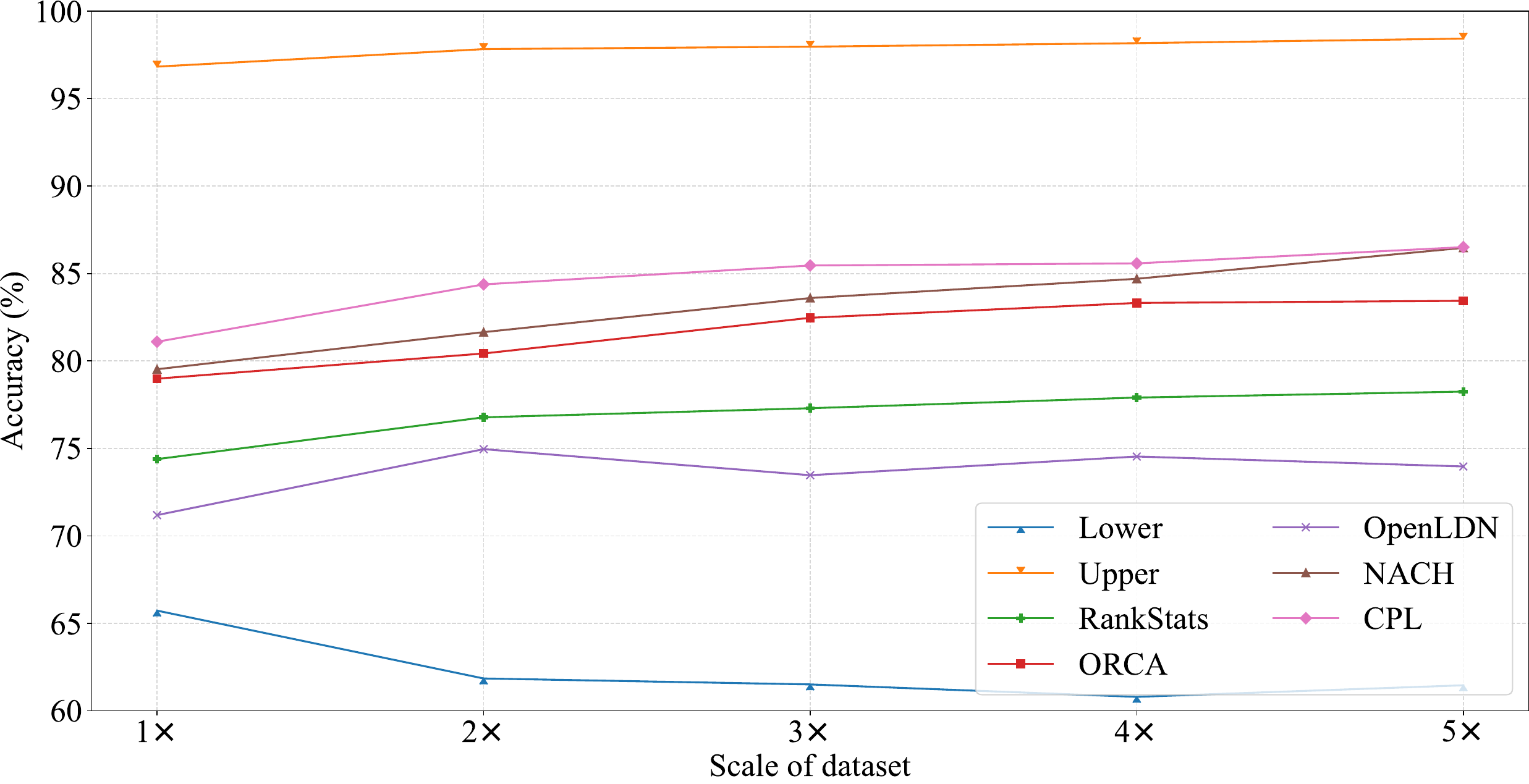}}
    \caption{Study of the relation between the performance of different methods and the scale of the dataset.}
    \label{fig:scale}
\end{figure*}

\begin{figure*}[t]
    \centering
    \includegraphics[width=0.82\linewidth]{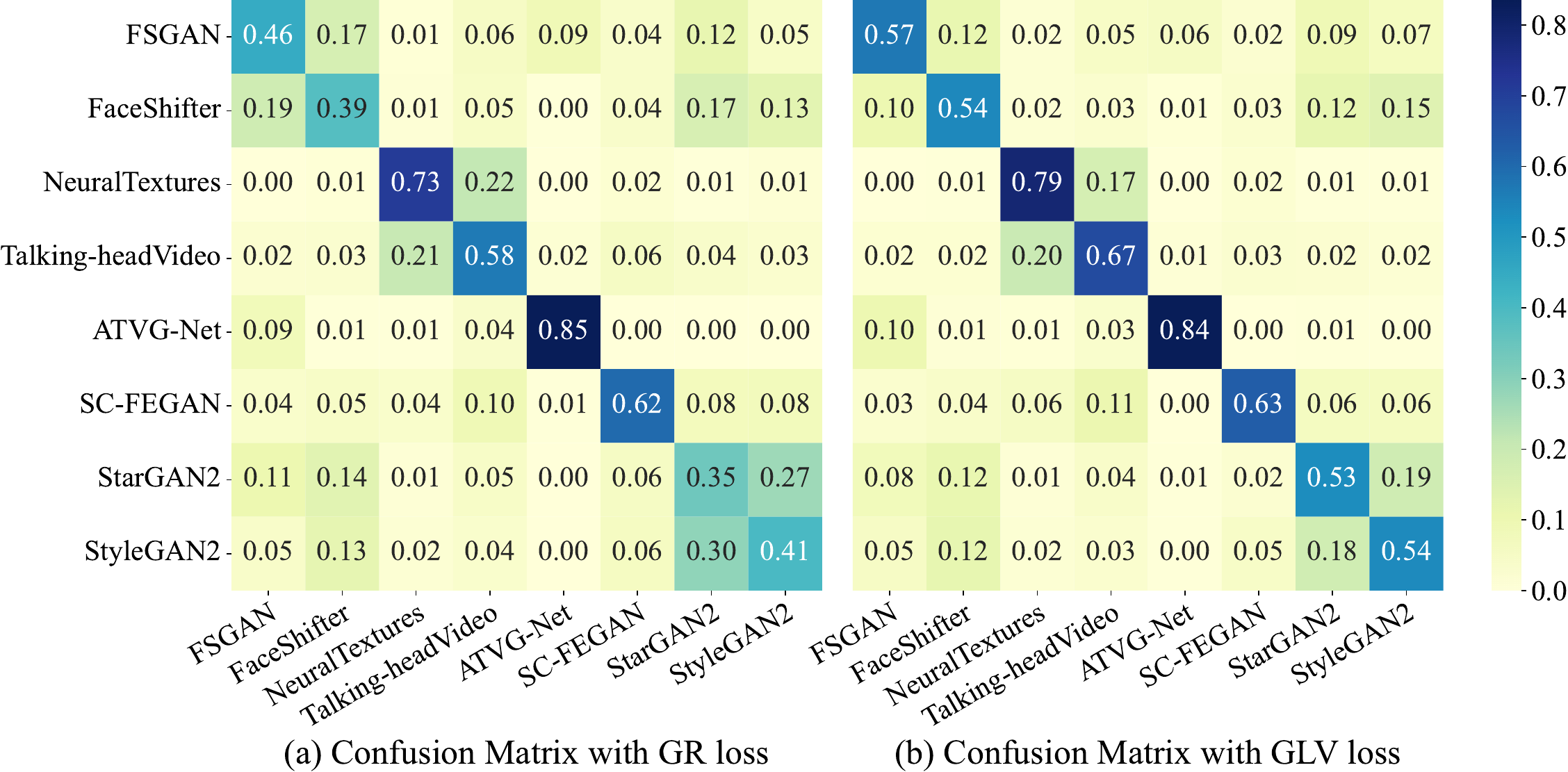}
    \caption{This confusion matrix displays the correct ratio of sample pairing using (a) GR loss and (b) GLV loss. The X-axis represents the actual forgery method, while the Y-axis represents the predicted forgery method.}
    \label{fig:confusion}
\end{figure*}

\begin{table}[t]
\vspace{0.3cm}
\centering
\caption{Ablation study of patch division on \textbf{Protocol-1}.}
\vspace{0.2cm}
\label{tab:patch}
\resizebox{\columnwidth}{!}{%
\begin{tabular}{cccccccc}
\toprule
\multicolumn{1}{c}{\multirow{2}{*}{\textbf{Patch}}} & \textbf{Known} & \multicolumn{3}{c}{\textbf{Novel}} & \multicolumn{3}{c}{\textbf{All}} \\ \cmidrule(lr){2-2}  \cmidrule(lr){3-5} \cmidrule(lr){6-8}
\multicolumn{1}{c}{} & \textbf{ACC} & \textbf{ACC} & \textbf{NMI} & \textbf{ARI} & \textbf{ACC} & \textbf{NMI} & \textbf{ARI} \\ \midrule
$3 \times 3$ & \textbf{97.50} & \textbf{71.89} & \textbf{68.20} & \textbf{59.37} & \textbf{83.70} & \textbf{82.31} & \textbf{77.64}  \\
$5 \times 5$ & 96.80 & 69.66 & 66.35 & 55.25 & 82.41 & 81.20 & 75.56 \\
$7 \times 7$ & 96.68 & 67.13 & 64.70 & 52.88 & 81.12 & 80.93 & 75.15 \\ \bottomrule
\end{tabular}%
}
\vspace{-0.1cm}
\end{table}

\section{Implementation for Experiments}
Due to space limitations in the main text, we have omitted some experimental details. In this section, we provide additional explanations for the specific implementation of those experiments.

\textbf{Implementation Details of the GLVM ablation study.}
To fairly compare the strengths and weaknesses of different methods in similarity learning, we compare the accuracy of similarity pair matching at various training stages. We use the ground truth of unlabeled samples to distinguish between known and novel classes. To visually represent this selection process, we calculate the accuracy of sample pairing and present it as a line chart. Further validation results for each forgery method can be found in Figure~\ref{fig:confusion}.

\textbf{Implementation Details of the PPLM ablation study.}
To ensure an equitable comparison between methods, we exclude strong and weak augmentation strategies due to their inapplicability to the OW-DFA task. Since the pseudo-label strategy relies on prior similarity learning, we use the GLV loss constraint as a baseline to ensure that the feature extractor and classifier have some ability to classify novel classes. In our comparison of all methods, we uniformly use a weight of $0.5$ for the pseudo-label cross-entropy loss.
Directly assigning labels refers to the strategy of choosing the prediction with the highest output value as the label. For the fixed-threshold approach, we use a threshold of $0.95$ for both known and novel classes and only assign labels to predictions that exceed this threshold. For dynamic-threshold approaches~\cite{zhang2021flexmatch,wang2022freematch}, we reproduce them using their open-source code and default configuration.
For ST Gumbel Softmax, we directly use the output of Gumbel Softmax as the label with a default temperature of $\tau = 1$.

\textbf{Implementation Details of Real/Fake Detection.}
To verify the importance of deepfake attribution for deepfake detection, we compare the performance of the deepfake detection task based on Protocol-2. We compare the results of three approaches: a) Deepfake binary classification, b) Deepfake multi-classification, and c) the CPL framework.
\textbf{a) Deepfake binary classification} is trained on the labeled set and outputs 0/1 to represent fake/real. When testing, we directly evaluate the performance based on the AUC result.
\textbf{b) Deepfake multi-classification} is trained on the labeled set with $9$ classifier outputs representing $1$ real face and $8$ forgery methods. Since there is only one real face type, we directly evaluate the AUC results using predicted output when testing.
\textbf{c) The CPL framework} is trained on both labeled and unlabeled sets using semi-supervised learning with $22$ classifier outputs representing $2$ real faces and $20$ forgery methods. Since multiple real face types appear, we first acquire the mapping relationship between prediction results and ground truth labels using the Hungarian algorithm~\cite{kuhn1955hungarian}. Then during testing, we sum up all prediction results for real faces to evaluate AUC results.

\vspace{0.1cm}

\section{Additional Experiments}

\textbf{Ablation Study on Scale of Dataset.}
To assess the scalability of each method further, we conduct an additional experiment to evaluate the performance of different methods on datasets of varying sizes. Due to the limited performance of DNA-Det~\cite{yang2022deepfake} and Openworld-GAN\cite{girish2021towards} in the OW-DFA task, we exclude these two methods from this experiment. Specifically, based on our original dataset
in Table~\ref{tab:dataset_real},
we scale up both Train and Test set to $2\times\!\sim\!5\times$ their original size.
The results of the ablation study are shown in Figure~\ref{fig:scale}. As expected, the performance of each method improves to some extent as the size of the dataset increases. However, our proposed method CPL consistently achieves the best results across all dataset sizes. We provide the specific settings for the $5\times$ scale of dataset in Table~\ref{tab:dataset_5x}, and the corresponding evaluation results are presented in Table~\ref{tab:compare5}.

\textbf{Confusion Matrix for Different Forgery Methods.} 
We record the predicted result and actual label of samples during similarity learning to analyze factors that contribute to ineffective classification. We present this information using a confusion matrix in Figure~\ref{fig:confusion}. To focus on categories with high confusion, we filter out all categories with prediction accuracy $>\!90\%$ and only include methods with low classification results. The method with GLR loss constraint can reduce confusion between similar categories while obtaining more accurate predictions. It has an accuracy of $>\!50\%$ on all categories. However, some samples are still confused with each other, especially when a) their data source is the same, such as StarGAN2, FaceShifter, and StyleGAN2, or when b) they belong to the same forgery type, including the confusion of NeuralTextures and Talking-Head-Video, and that of FaceShifter and FSGAN.

\textbf{Ablation Study on Patch Division.} 
To compare the performance of different patch sizes and evaluate their impact on overall performance, we conduct an ablation study on patch division. Table~\ref{tab:patch} presents the results of the ablation study, which show that the optimal performance is achieved with a smaller number of patch splits of $3 \times 3$. Specifically, we observe that using a smaller grid for local region partitioning can alleviate the problem of the same forged region being sliced into different local patches.

\end{document}